\documentclass[10pt,twocolumn,letterpaper]{article}

\usepackage[pagenumbers]{cvpr} %

\usepackage{graphicx}
\usepackage{amsmath}
\usepackage{amssymb}
\usepackage{booktabs}
\usepackage{mathtools}
\usepackage{multirow}
\usepackage{cuted}

\usepackage{amssymb}%
\usepackage{pifont}%
\newcommand{\cmark}{\ding{51}}%
\newcommand{\xmark}{\ding{55}}%

\usepackage[pagebackref,breaklinks,colorlinks]{hyperref}

\usepackage[capitalize]{cleveref}
\crefname{section}{Sec.}{Secs.}
\Crefname{section}{Section}{Sections}
\Crefname{table}{Table}{Tables}
\crefname{table}{Tab.}{Tabs.}

\def\cvprPaperID{1020} %
\def\confName{CVPR}
\def\confYear{2023}

\begin{document}

\title{Behind the Scenes: Density Fields for Single View Reconstruction}

\author{Felix Wimbauer$^{1,2}$ \hspace{1cm} Nan Yang$^1$ \hspace{1cm} Christian Rupprecht$^3$ \hspace{1cm} Daniel Cremers$^{1,2,3}$\\
$^1$Technical University of Munich \hspace{1cm} $^2$MCML \hspace{1cm} $^3$University of Oxford\\
{\tt\small \{felix.wimbauer, nan.yang, cremers\}@tum.de \hspace{.5cm} chrisr@robots.ox.ac.uk}
}
\maketitle

\begin{strip}
\vspace{-1.5cm}
\centering
\captionsetup{type=figure}
\includegraphics[trim={0cm .15cm 0cm 0cm},clip,width=\linewidth]{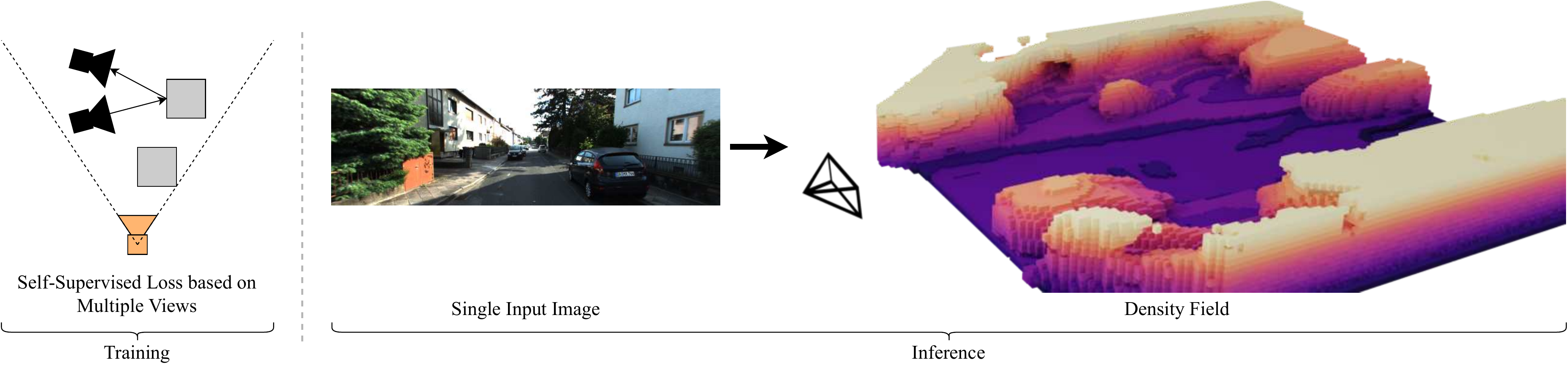}
\vspace{-.7cm}
\captionof{figure}{\textbf{Predicting a Density Field from a Single Image.} Through a novel ``density field'' formulation, which decouples geometry from color, architectural improvements, and a novel self-supervised training scheme, our method learns to predict a volumetric scene representation from a single image in challenging conditions. The voxel occupancy view shows that our method predicts accurate density even in occluded regions, which is not possible in traditional depth prediction. 
Please check out our project page at \href{https://fwmb.github.io/bts/}{fwmb.github.io/bts/}.
}
\vspace{-.2cm}
\label{fig:teaser}
\end{strip}
\begin{abstract}
Inferring a meaningful geometric scene representation from a single image is a fundamental problem in computer vision.
Approaches based on traditional depth map prediction can only reason about areas that are visible in the image. %
Currently, neural radiance fields (NeRFs) can capture true 3D including color, but are too complex to be generated from a single image. %
As an alternative, we propose to predict implicit density fields.
A density field maps every location in the frustum of the input image to volumetric density.
By directly sampling color from the available views instead of storing color in the density field, our scene representation becomes significantly less complex compared to NeRFs, and a neural network can predict it in a single forward pass. 
The prediction network is trained through self-supervision from only video data.
Our formulation allows volume rendering to perform both depth prediction and novel view synthesis.
Through experiments, we show that our method is able to predict meaningful geometry for regions that are occluded in the input image.
Additionally, we demonstrate the potential of our approach on three datasets for depth prediction and novel-view synthesis.
\vspace{-.3cm}
\end{abstract}

\section{Introduction}
\label{sec:intro}

The ability to infer information about the geometric structure of a scene from a single image is of high importance for a wide range of applications from robotics to augmented reality. While traditional computer vision mainly focused on reconstruction from multiple images, in the deep learning age the challenge of inferring a 3D scene from merely a single image has received renewed attention.

Traditionally, this problem has been formulated as the task of predicting per-pixel depth values (\ie depth maps).
One of the most influential lines of work showed that it is possible to train neural networks for accurate single-image depth prediction in a self-supervised way only from video sequences. \cite{zhou2017unsupervised, zhan2018unsupervised, godard2019digging, luo2019every, guizilini20203d, shu2020feature, yuan2022new, gonzalezbello2020forget, watson2019self}
Despite these advances, depth prediction methods are \textit{not} modeling the true 3D of the scene: they model only a \textit{single} depth value per pixel.
As a result, it is not directly possible to obtain depth values from views other than the input view without considering interpolation and occlusion.
Further, the predicted geometric representation of the scenes does not allow reasoning about areas that lie \textit{behind} another object in the image (\eg a house behind a tree), inhibiting the applicability of monocular depth estimation to 3D understanding.

Due to the recent advance of 3D neural fields, the related task of novel view synthesis has also seen a lot of progress.
Instead of directly reasoning about the scene geometry, the goal here is to infer a representation that allows rendering views of the scene from novel viewpoints. While geometric properties can often be inferred from the representation, they are usually only a side product and lack visual quality.

Even though neural radiance field \cite{mildenhall2021nerf} based methods achieve impressive results, they require many training images per scene and do not generalize to new scenes. %
To enable generalization, efforts have been made to condition the neural network on global or local scene features.
However, this has only been shown to work well on simple scenes, for example, scenes containing an object from a single category \cite{yu2021pixelnerf, sharma2022seeing}.
Nevertheless, obtaining a neural radiance field from a single image has not been achieved before.

In this work, we tackle the problem of inferring a geometric representation from a single image by generalizing the depth prediction formulation to a continuous density field.
Concretely, our architecture contains an encoder-decoder network that predicts a dense feature map from the input image.
This feature map locally conditions a density field inside the camera frustum, which can be evaluated at any spatial point through a multi-layer perceptron (MLP).
The MLP is fed with the coordinates of the point and the feature sampled from the predicted feature map by reprojecting points into the camera view.
To train our method, we rely on simple image reconstruction losses.

Our method achieves robust generalization and accurate geometry prediction even in very challenging outdoor scenes through three key novelties:

\textbf{1. Color sampling.} 
When performing volume rendering, we sample color values directly from the input frames through reprojection instead of using the MLP to predict color values.
We find that only predicting density drastically reduces the complexity of the function the network has to learn.
Further, it forces the model to adhere to the multi-view consistency assumption during training, leading to more accurate geometry predictions.

\textbf{2. Shifting capacity to the feature extractor.} 
In many previous works, an \textit{encoder} extracts image features to condition local appearance, while a high-capacity MLP is expected to generalize to multiple scenes.
However, on complex and diverse datasets, the training signal is too noisy for the MLP to learn meaningful priors.
To enable robust training, we significantly reduce the capacity of the MLP and use a more powerful \textit{encoder-decoder} that can capture the entire scene in the extracted features.
The MLP then only evaluates those features locally.

\textbf{3. \textit{Behind the Scenes} loss formulation.}
The continuous nature of density fields and color sampling allow us to reconstruct a novel view from the colors of any frame, not just the input frame.
By applying a reconstruction loss between two frames that both observe areas occluded in the input frame, we train our model to predict meaningful geometry \textit{everywhere} in the camera frustum, not just the visible areas.

We demonstrate the potential of our new approach in a number of experiments on different datasets regarding the aspects of capturing true 3D, depth estimation, and novel view synthesis.
On KITTI \cite{geiger2013vision} and KITTI-360 \cite{liao2022kitti}, we show both qualitatively and quantitatively that our model can indeed capture true 3D, and that our model achieves state-of-the-art depth estimation accuracy.
On RealEstate10K \cite{silberman2012indoor} and KITTI, we achieve competitive novel view synthesis results, even though our method is purely geometry-based.
Further, we perform thorough ablation studies to highlight the impact of our design choices.

\section{Related Work}
\label{sec:related_work}

In the following, we review the most relevant works that are related to our proposed method.

\subsection{Single-Image Depth Prediction}

One of the predominant formulations to capture the geometric structure of a scene from a single image is predicting a per-pixel depth map.
Learning-based methods have proven able to overcome the inherent ambiguities of this task by correlating contextual cues extracted from the image with certain depth values.
One of the most common ways to train a method for single-image depth prediction is to immediately regress the per-pixel ground-truth depth values \cite{eigen2014depth, liu2015learning}.
Later approaches supplemented the fully-supervised training with reconstruction losses \cite{kuznietsov2017semi, yang2018deep}, or specialise the architecture and loss formulation \cite{fu2018deep, li2022binsformer, li2022depthformer, lee2021patch, aich2021bidirectional, wimbauer2021monorec}.
To overcome the need for ground-truth depth annotations, several papers focused on relying exclusively on reconstruction losses to train prediction networks.
Both temporal video frames \cite{zhou2017unsupervised} and stereo frames \cite{godard2017unsupervised}, as well as combinations of both \cite{zhan2018unsupervised, godard2019digging} can be used as the reconstruction target.
Different follow-up works refine the architecture and loss \cite{luo2019every, guizilini20203d, shu2020feature, yuan2022new, gonzalezbello2020forget, watson2019self}.
\cite{zhou2022devnet} first predicts a discrete density volume as an intermediate step, from which depth maps can be rendered from different views.
While they use this density volume for regularization, their focus is on improving depth prediction and their method does not demonstrate the ability to learn true 3D.

\subsection{Neural Radiance Fields}

Many works have investigated alternative approaches to representing scenes captured from a single or multiple images, oftentimes with the goal of novel view synthesis.
Recently, \cite{mildenhall2021nerf} proposed to represent scenes as neural radiance fields (NeRFs). 
In NeRFs, a multi-layer perceptron (MLP) is optimized per scene to map spatial coordinates to color (appearance) and density (geometry) values.
By evaluating the optimized MLP along rays and then integrating the color over the densities, novel views can be rendered under the volume rendering formulation \cite{max1995optical}.
Training data consists of a large number of images of the same scene from different viewpoints with poses computed from traditional SFM and SLAM methods \cite{schoenberger2016mvs, schoenberger2016sfm, campos2021orb}.
The training goal is to reconstruct these images as accurately as possible.
NeRF's impressive performance inspired many follow-up works, which improve different parts of the architecture \cite{barron2021mip, barron2022mip, niemeyer2022regnerf, kim2022infonerf, jain2021putting, deng2022depth, roessle2022dense}.

In the traditional NeRF formulation, an entire scene is captured in a single, large MLP.
Thus, the trained network cannot be adapted to different settings or used for other scenes.
Further, the MLP has to have a high capacity, resulting in slow inference.
Several methods propose to condition such MLPs on feature grids or voxels \cite{mescheder2019occupancy, peng2020convolutional, liu2020neural, takikawa2021neural, chen2022tensorf, muller2022instant, yu2021pixelnerf, sharma2022seeing}.
Through this, the MLP needs to store less information and can be simplified, speeding up inference \cite{liu2020neural, takikawa2021neural, chen2022tensorf, muller2022instant}.
Additionally, this allows for some generalization to new scenes \cite{yu2021pixelnerf, sharma2022seeing, muller2022autorf}. 
However, generalization is mostly limited to a single object category, or simple synthetic data, where the scenes differ in local details.
In contrast, our proposed method can generalize to highly complex outdoor scenes.
\cite{cao2022scenerf} also improves generalization through depth supervision and improved ray sampling.

\subsection{Single Image Novel View Synthesis}

While traditional NeRF-based methods achieve impressive performance when provided with enough images per scene, they do not work with only a single image of a scene available. 
In recent years, a number of methods for novel-view synthesis (NVS) from a single image emerged.

Several methods \cite{tulsiani2018layer, dhamo2019object, dhamo2019peeking} predict layered depth images (LDI) \cite{shade1998layered} for rendering.
Later approaches \cite{tucker2020single, srinivasan2019pushing}  directly produce a multiplane image (MPI) \cite{zhou2018stereo}. %
\cite{li2021mine} predicts a generalized multiplane image. 
Instead of directly outputting the discrete layers, the architecture's decoder receives a variable depth value, for which it outputs the layer.
In \cite{wiles2020synsin}, a network predicts both a per-pixel depth and feature map, which are then used in a neural rendering framework.
Other works \cite{wimbauer2022rendering, wu2020unsupervised} perform image decomposition, followed by classical rendering.
While these methods achieve impressive NVS results, the quality of predicted geometry usually falls short.
Some methods even predict novel views without any geometric representation \cite{zhou2016view, sajjadi2022scene}.

\section{Method}

\begin{figure*}
    \centering
    \includegraphics[width=\linewidth]{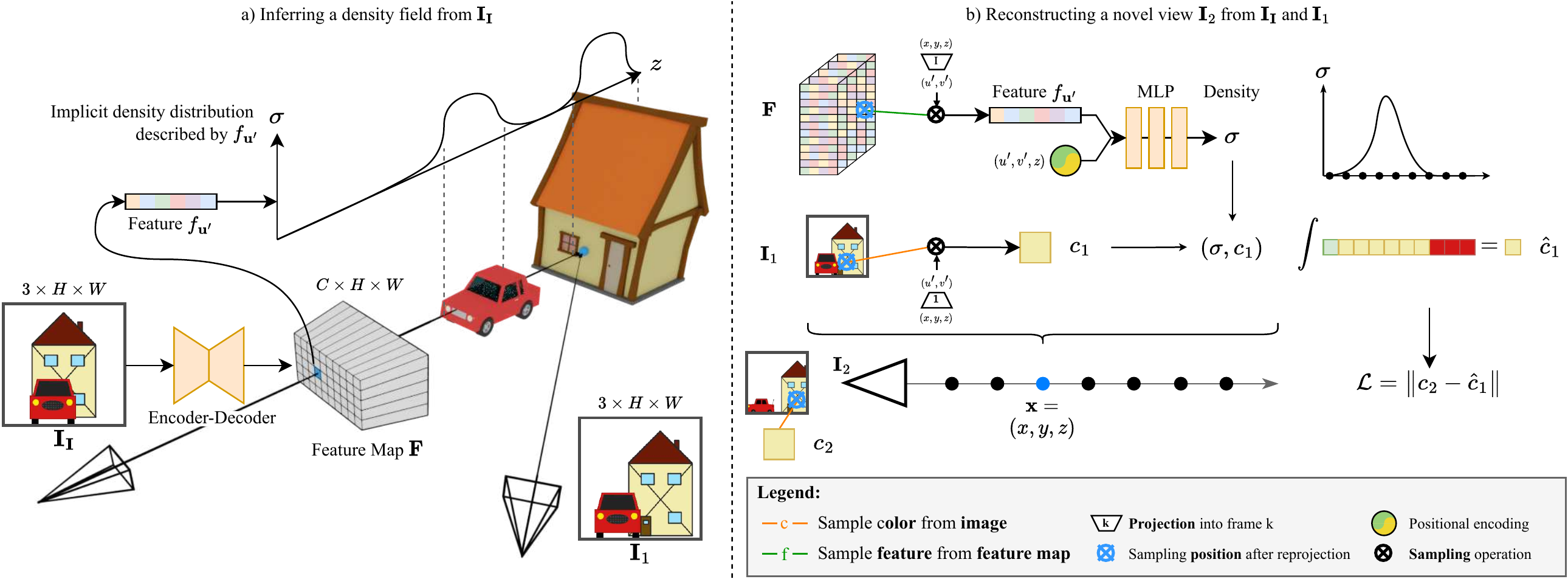}
    \caption{\textbf{Overview.} \textit{a)} Our method first predicts a pixel-aligned feature map $\textbf{F}$, which describes a density field, from the input image $\textbf{I}_\textbf{I}$. For every pixel $\textbf{u}^\prime$, the feature $f_{\textbf{u}^\prime}$ implicitly describes the density distribution along the ray from the camera origin through $\textbf{u}^\prime$. Crucially, this distribution can model density even in occluded regions (\eg the house). \textit{b)} To render novel views, we perform volume rendering. For any point $\textbf{x}$, we project $\textbf{x}$ into $\textbf{F}$ and sample $f_{\textbf{u}^\prime}$. This feature is combined with positional encoding and fed into an MLP to obtain density $\sigma$. We obtain the color $c$ by projecting $\textbf{x}$ into one of the views, in this case, $\textbf{I}_1$, and directly sampling the image. }
    \label{fig:architecture}
    \vspace{-.2cm}
\end{figure*}

In the following, we describe a neural network architecture that predicts the geometric structure of a scene from a single image $\textbf{I}_\text{I}$, as shown in \cref{fig:architecture}.
We first cover how we represent a scene as a continuous density field, and then propose a training scheme that allows our architecture to learn geometry even in occluded areas.

\subsection{Notation}

Let $\textbf{I}_\text{I} \in [0, 1]^{3\times H\times W} = (\mathbb{R}^3)^\Omega$ be the input image, defined on a lattice $\Omega = \{1, \ldots, H\}\times\{1, \ldots, W\}$.
$T_\text{I} \in \mathbb{R}^{4\times 4}$ and $K_\text{I} \in \mathbb{R}^{3 \times 4}$ are the corresponding world-to-camera pose matrix and projection matrix, respectively.
During training, we have available an additional set of $N = \{1, 2, \dots, n\}$ frames $\textbf{I}_k, k\in N$ with corresponding world-to-camera pose and projection matrices $T_k, K_k, k\in N$.
When assuming homogeneous coordinates, a point $\textbf{x} \in \mathbb{R}^3$ in world coordinates can be projected onto the image plane of frame $k$ with the following operation: $\pi_k(\textbf{x}) = K_k T_k \textbf{x}$

\subsection{Predicting a Density Field}

We represent the geometric structure of a scene as a function, which maps scene coordinates $\textbf{x}$ to volume density $\sigma$.
We term this function "density field".
Inference happens in two steps.
From the input image $\textbf{I}_\text{I}$, an encoder-decoder network first predicts a pixel-aligned feature map $\textbf{F} \in (\mathbb{R}^C)^\Omega$.
The idea behind this is that every feature $f_{\textbf{u}} = \textbf{F}(\textbf{u})$ at pixel location $\textbf{u} \in \Omega$ captures the distribution of local geometry along the ray from the camera origin through the pixel at $\textbf{u}$.
It also means that the density field is designed to lie inside the camera frustum. 
For points outside of this frustum, we extrapolate features from within the frustum.

To obtain a density value at a 3D coordinate $\textbf{x}$, we first project $\textbf{x}$ onto the input image $\textbf{u}^\prime_\text{I} = \pi_{\text{I}}(\textbf{x})$ and bilinearly sample the feature $f_{\textbf{u}^\prime} = \textbf{F}(\textbf{u}^\prime)$ at that position. 
This feature $f_{\textbf{u}^\prime}$, along with the positional encoding \cite{mildenhall2021nerf} $\gamma(d)$ of the distance $d$ between $\textbf{x}$ and the camera origin, and the positional encoding $\gamma(\textbf{u}^\prime_\text{I})$ of the pixel, is then passed to a multi-layer perceptron (MLP) $\phi$.
During training, $\phi$ and $\textbf{F}$ learn to describe the density of the scene given the input view. 
We can interpret the feature representation $f_{\textbf{u}^\prime}$ as a descriptor of the density along a ray through the camera center and pixel $\textbf{u}^\prime$.
In turn, $\phi$ acts as a decoder, that given $f_{\textbf{u}^\prime}$ and a distance to the camera, predicts the density at the 3D location $\textbf{x}$.
\begin{equation}
    \sigma_\textbf{x} = \phi(f_{\textbf{u}^\prime_\text{I}}, \gamma(d), \gamma(\textbf{u}^\prime_\text{I}))
\end{equation}
Unlike most current works on neural fields, we \textit{do not} use $\phi$ to also predict color.
This drastically reduces the complexity of the distribution along a ray as density distributions tend to be simple, while color often contains complex high-frequency components.
In our experiments, this makes capturing such a distribution in a single feature, so that it can be evaluated by an MLP, %
much more tractable.

\subsection{Volume Rendering with Color Sampling}
When rendering the scene from a novel viewpoint, we do not retrieve color from our scene representation directly. 
Instead, we sample the color for a point in 3D space from the available images. 
Concretely, we first project a point $\textbf{x}$ into a frame $k$ and then bilinearly sample the color $ c_{\textbf{x}, k} = \textbf{I}_k (\pi_k(\textbf{x})).$

By combining $\sigma_\textbf{x}$ and $c_{\textbf{x}, k}$, we can perform volume rendering \cite{kajiya1984ray, max1995optical} to synthesize novel views.
We follow the discretization strategy of other radiance field-based methods, \eg \cite{mildenhall2021nerf}.
To obtain the color $\hat{c}_k$ for a pixel in a novel view, we emit a ray from the camera and integrate the color along the ray over the probability of the ray ending at a certain distance.
To approximate this integral, density and color are evaluated at $S$ discrete steps $\textbf{x}_i$ along the ray. 
Let $\delta_i$ be the distance between $\textbf{x}_i$ and $\textbf{x}_{i+1}$, and $\alpha_i$ be the probability of a ray ending between $\textbf{x}_i$ and $\textbf{x}_{i+1}$.
From the previous $\alpha_j$s, we can compute the probability $T_i$ that the ray does not terminate before $\textbf{x}_i$, \ie the probability that $\textbf{x}_i$ is not occluded.
\begin{equation}
    \alpha_i = \exp(1 - \sigma_{\textbf{x}_i} \delta_i)\quad\quad
    T_i = \prod_{j=1}^{i-1} (1 - \alpha_j)%
\end{equation}
\begin{equation}
    \hat{c}_k = \sum_{i=1}^{S}T_i \alpha_i c_{\textbf{x}_i, k}
\end{equation}
Similarly, we can also retrieve the expected ray termination depth, which corresponds to the depth in a depth map.
Let $d_i$ be the distance between $\textbf{x}_i$ and the ray origin.
\begin{equation}
    \hat{d} = \sum_{i=1}^{S}T_i \alpha_i d_i
\end{equation}

This rendering formulation is very flexible.
We can sample the color values from any frame, and, crucially, it can be a different frame from the input frame.
It is even possible to obtain multiple colors from multiple different frames for a single ray, which enables reasoning about occluded areas during training.
Note that even though different frames can be used, the density is always based on features from the input image and does not change.
During inference, color sampling from different frames is not necessary, everything can be done based on a single input image.

\subsection{\textbf{\textit{Behind the Scenes}} Loss Formulation}

Our training goal is to optimize both the encoder-decoder network and $\phi$ to predict a density field only from the input image, such that it allows the reconstruction of other views.

Similar to radiance fields and self-supervised depth prediction methods, we rely on an image reconstruction loss.
For a single sample, we first compute the feature map $\textbf{F}$ from $\textbf{I}_\text{I}$ and randomly partition \textit{all} frames $\hat{N} = \{\text{I}_\text{I}\} \cup N$ into two sets $N_\text{loss}, N_\text{render}$. 
Note that the input image can end up in any of the two sets.
We reconstruct the frames in $N_\text{loss}$ by sampling colors from $N_\text{render}$ using the camera poses and the predicted densities. 
The photometric consistency between the reconstructed frames and the frames in $N_\text{loss}$ serves as supervision for the density field. 
In practice, we randomly sample $p$ patches $P_i$ to use patch-wise photometric measurement.
For every patch $P_i$ in $N_\text{loss}$, we obtain a reconstructed patch $\hat{P}_{i, k}$ from \textit{every} frame $k \in N_\text{render}$. We aggregate the costs between $P_i$ and every $\hat{P}_{i, k}$ by taking the per-pixel \textit{minimum} across the different frames $k$, similar to \cite{godard2019digging}. The intuition behind this is that for every patch, there is a frame in $N_\text{render}$, which ``sees'' the same surface.
Therefore, if the predicted density is correct, then it results in a very good reconstruction and a low error.
\begin{figure}
    \centering
    \includegraphics[width=.8\linewidth]{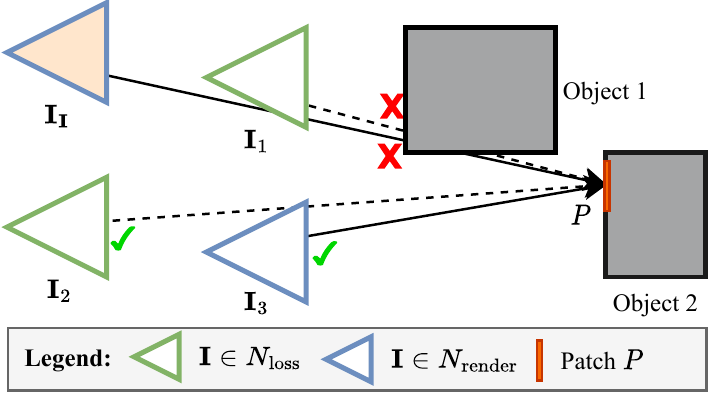}
    \caption{\textbf{Loss in Occluded Regions.} Patch $P$ on Object 2 is occluded by Object 1 in the input frame $\textbf{I}_\textbf{I}$ and $\textbf{I}_1$. In order to correctly reconstruct $P$ in $\textbf{I}_2$ from $\textbf{I}_3$, the network needs to predict density for Object 2 \textit{behind} Object 1.\label{fig:loss}}
    \vspace{-.3cm}
\end{figure}

For the final loss formula, we use a combination of L1 and SSIM \cite{wang2004image} to compute the photometric discrepancy, as well as an edge-aware smoothness term.
Let $d^\star_i$ denote the inverse, mean-normalized expected ray termination depth of patch $P_i$.
Both $\mathcal{L}_\text{ph}$ and $\mathcal{L}_\text{eas}$ are computed per $(x, y)$ element of the patch, thus resulting in 2D loss maps.
They are then aggregated when computing $\mathcal{L}$.

\begin{equation}
    \mathcal{L}_\text{ph} = \min_{k\in N_{\text{render}}} \left(\lambda_\text{L1} \text{L1}(P_i, \hat{P}_{i, k}) + \lambda_\text{SSIM} \text{SSIM}(P_i, \hat{P}_{i, k})\right)
\label{eq:loss_ph}
\end{equation}
\begin{equation}
    \mathcal{L}_\text{eas} = \left|\delta_x d^\star_i\right|e^{-\left|\delta_x P_i\right|} + \left|\delta_y d^\star_i\right|e^{-\left|\delta_y P_i\right|}
\label{eq:loss_eas}
\end{equation}
\begin{equation}
    \mathcal{L} = \sum_{i=1}^{p} \sum_{x, y \in P} \left(\mathcal{L}_\text{ph} + \lambda_\text{eas}\mathcal{L}_\text{eas} \right)(x, y)
\label{eq:loss}
\end{equation}
\vspace{-.5cm}

\paragraph{Learning true 3D.}
Our loss formula \cref{eq:loss} is the same as for self-supervised depth prediction methods, like \cite{godard2019digging}.
The key difference, however, is that depth prediction methods can only densely reconstruct the input image, for which the per-pixel depth was predicted.

In contrast, our density field formulation allows us to reconstruct \textit{any} frame from \textit{any other} frame.
Consider an area of the scene, which is occluded in the input $\textbf{I}_\text{I}$, but visible in two other frames $\textbf{I}_\text{k}, \textbf{I}_\text{k+1}$, as depicted in \cref{fig:loss}:
During training, we aim to reconstruct this area in $\textbf{I}_\text{k}$.
The reconstruction based on colors sampled from $\textbf{I}_\text{k+1}$ will give a clear training signal to correctly predict the geometric structure of this area, even though it is occluded in $\textbf{I}_\text{I}$.
Note, that in order to learn geometry about occluded areas, we require at least \textbf{two additional} views besides the input during training, \ie to look \textit{behind the scenes}.

\paragraph{Handling invalid samples.}
While the frustums of the different views overlap for the most part, there is still a chance of a ray leaving the frustums, thus sampling invalid features, or sampling invalid colors.
Such invalid rays lead to noise and instability in the training process.
Therefore, we propose a policy to detect and remove invalid rays.
Our intuition is that when the amount of contribution to the final aggregated color, that comes from invalidly sampled colors or features, exceeds a certain threshold $\tau$, the ray should be discarded.
Consider a ray that is evaluated at positions $\textbf{x}_i, i\in [1, 2, \dots, S]$ and reconstructed from frames $K$:
$O_{i, k}, k \in \{\text{I}\} \cup K$ denotes the indicator function that $\textbf{x}_i$ is outside the camera frustum of frame $k$.
Note that we always sample features from the input frame.
We define $\text{IV}(k)$ to be the function indicating that the rendered color based on frame $k$ is invalid as:
\begin{equation}
    \text{IV}(k) = \sum_{i=1}^{S}T_i \alpha_i \left(O_{i, \text{I}} \vee O_{i, k}\right) > \tau
\end{equation}
Only if $\text{IV}(k)$ is true for \textit{all} frames the ray was reconstructed from, we ignore the ray when computing the loss value.
The reasoning behind this is that non-invalid rays will still lead to the lowest error.
Therefore, the $\min$ operation in \cref{eq:loss_ph} will ignore the invalid rays.

\subsection{Implementation Details}

We implement our model in PyTorch \cite{Paszke_PyTorch_An_Imperative_2019} on a single Nvidia RTX A40 GPU with 48GB memory. 
The encoder-decoder network follows \cite{godard2019digging} using a ResNet encoder \cite{he2016deep} and predicts feature maps with 64 channels.
The MLP $\phi$ is made lightweight with only 2 fully connected layers and 64 hidden nodes each. 
We use a batch size of 16 and sample 32 patches of size $8\times 8$ from the images for which we want to compute the reconstruction loss.
Every ray is sampled at 64 locations, based on linear spacing in inverse depth.
For more details, \eg exact network architecture and further hyperparameters, please refer to the supplementary material.

\section{Experiments}

To demonstrate the abilities and advantages of our proposed method, we conduct a wide range of experiments.
First, we demonstrate that our method is uniquely able to capture a holistic geometry representation of the scene, even in areas that are occluded in the input image.
Additionally, we also show the effect of different data setups on the prediction quality.
Second, we show that our method, even though depth maps are only a side product of our scene representation, achieves depth accuracy on par with other state-of-the-art self-supervised methods, that are specifically designed for depth prediction.
Third, we demonstrate that, even though our representation is geometry-only, our method can be used to perform high-quality novel view synthesis from a single image.
Finally, we conduct thorough ablation studies based on occupancy estimation and depth prediction to justify our design choices.

\subsection{Data}

\begin{figure*}
    \centering
    \includegraphics[trim={0cm .35cm 0cm 0cm},width=.9\linewidth]{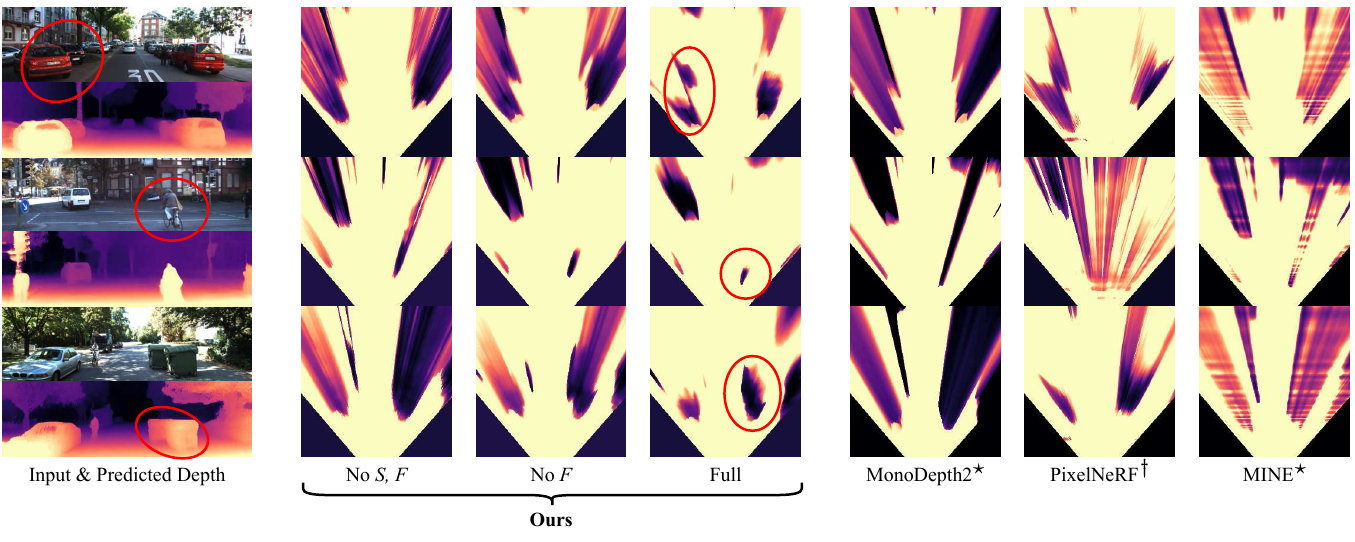}
    \caption{\textbf{Occupancy Estimation.} Top-down visualization of predicted occupancy volumes. We show an area of $x=[-9m, 9m], z=[3m, 21m]$ %
    and $y=[0m,1m]$ (just above the road).
    Our method produces an accurate volumetric reconstruction, even for occluded regions. 
    Training with more views improves the quality.
    Depth prediction methods like MonoDepth2 \cite{godard2019digging} do not predict a full 3D volume. 
    Thus, objects cast ``occupancy shadows'' behind them.  
    Volumetric methods like PixelNeRF \cite{yu2021pixelnerf} and MINE \cite{li2021mine} produce noisy predictions.
    Inference is from a single image.
    \textbf{Legend}: \textit{S}: Stereo, \textit{F}: Fisheye, $\star$: official checkpoint, $\dagger$: trained in same setup as our Full variant.
    \label{fig:profile}}
    \vspace{-.2cm}
\end{figure*}
For our experiments, we use three different datasets: KITTI \cite{geiger2013vision}, KITTI-360 \cite{liao2022kitti} (autonomous driving), and RealEstate10K \cite{zhou2018stereo} (indoor).
RealEstate10K only has monocular sequences, while KITTI and KITTI-360 provide stereo. 
KITTI-360 also contains fisheye camera frames facing left and right.
For monocular data, we use three timesteps per sample, for stereo sequences (possibly with fisheye frames), we use two timesteps.
The fisheye frames are offset by one second to increase the overlap of the different camera frustums.%
\footnote{More details on offsets, pose data, and data splits in the supp.\ mat.}
Training is performed for 50 epochs on KITTI (approx.\ 125k steps), 25 epochs on KITTI-360 (approx.\ 150k steps), and 360k iterations on RealEstate10K.
We use a resolution of $640\times 192$ for KITTI and KITTI-360, and follow \cite{li2021mine} in using a resolution of $384\times 256$ for RealEstate10K.

\subsection{Capturing true 3D}

\begin{table}[]
\centering
\footnotesize
\begin{tabular}{lccc}
\toprule
\textit{Method} & $\text{O}_\text{acc} \uparrow$ & $\text{IE}_\text{acc} \uparrow$ & $\text{IE}_\text{rec} \uparrow$ \\
\midrule
Depth$^\dagger$ \cite{godard2019digging}     & \textbf{0.94}   & n/a         & n/a         \\
Depth$^\dagger$ \textit{+ 4m} \cite{godard2019digging} & 0.91   & 0.63        & \underline{0.22}        \\
PixelNeRF$^\dagger$ \cite{yu2021pixelnerf}  & \underline{0.92}   & 0.63        & \textbf{0.43}        \\
\midrule
\textbf{Ours} (No \textit{S, F})       & \textbf{0.94}   & 0.70        & 0.06        \\
\textbf{Ours} (No \textit{F})     & \textbf{0.94}   & \underline{0.71}        & 0.09        \\
\textbf{Ours} & \textbf{0.94}   & \textbf{0.77}        & \textbf{0.43}        \\
\bottomrule
\end{tabular}
\caption{\textbf{3D Scene Occupancy Accuracy on KITTI-360.} We evaluate the capability of the model to predict occupancy \textit{behind} objects in the image. 
Ground truth occupancy maps are computed from 20 consecutive Lidar scans per frame.
Depth prediction \cite{godard2019digging} naturally has no ability to predict behind occlusions.
PixelNeRF \cite{yu2021pixelnerf} can predict free space in occluded regions, but produces poor overall geometry. Our method improves when training with more views. Inference from a single image. Samples are evenly spaced in a cuboid $w=[-4m, 4m], h=[-1m, 0m], d=[3m, 20m]$ relative to the camera. \textbf{Legend}: ref.\ \cref{fig:profile}.\label{tab:lidar_occ}}%
\vspace{-.2cm}
\end{table}
Evaluation of fully geometric 3D representations like density fields is difficult.
Real-world datasets usually only provide ground truth data captured from a single viewpoint, \eg RGB-D frames and Lidar measurements.
Nevertheless, we aim to evaluate and compare this key advantage of our method both qualitatively and quantitatively.
Through our proposed training scheme, our networks are able to learn to also predict meaningful geometry in occluded areas. %

To overcome the lack of volumetric ground truth, we accumulate Lidar scans to build reference occupancy maps for KITTI-360.
Consider a single input frame for which we want to evaluate an occupancy prediction:
As KITTI-360 is a driving dataset with a forward-moving camera, the consecutive Lidar scans captured a short time later measure different areas within the camera frustum.
Note that these Lidar measurements can reach areas that are occluded in the input image.
To determine whether a point is occupied, we check whether it is \textit{in front} of the measured surface for any of the Lidar scans.
Intuitively, every Lidar measurement ``carves out'' unoccupied areas in 3D space.
By accumulating enough Lidar scans, we obtain a reliable occupancy measurement of the entire camera frustum.
Whether a point is visible in the input frame can be checked using the Lidar scan corresponding to the input frame.\footnote{More details on the exact procedure and examples in the supp.\ mat.}

For every frame, we sample points in a cuboid area in the camera frustum and compute the following metrics: 1.\ Occupancy accuracy ($\text{O}_\text{Acc}$), 2.\ Invisible and empty accuracy ($\text{IE}_\text{Acc}$), and 3.\ Invisible and empty recall ($\text{IE}_\text{Rec}$).
$\text{O}_\text{Acc}$ evaluates the occupancy predictions across the whole scene volume.
$\text{IE}_\text{Acc}$ and $\text{IE}_\text{Rec}$ specifically evaluate invisible regions, evaluating performance beyond depth prediction.

We train a MonoDepth2 \cite{godard2019digging} model to serve as a baseline representing ordinary depth prediction methods.
Here, we consider all points behind the predicted depth to be occupied.
Additionally, we evaluate a version in which we consider points only up to 4 meters (average car length) behind the predicted depth as occupied.
As a second baseline, we train a PixelNeRF \cite{yu2021pixelnerf} model, one of the most prominent NeRF variants that also has the ability to generalize.

To demonstrate that our loss formulation generates strong training signals for occluded regions, given the right data, we train our model in several different data configurations.
By removing the fisheye, respectively fisheye, and stereo frames, the training signal for occluded areas becomes much weaker.
\cref{tab:lidar_occ} reports the obtained results.

The depth prediction baselines achieve a strong overall accuracy, but are, by design, not able to predict meaningful free space in occluded areas.
PixelNeRF can predict free space in occluded areas but produces poor overall geometry.
Our model achieves strong overall accuracy, while it is also able to recover the geometry of the occluded parts of the scene.
Importantly, our model becomes better at predicting \textit{free space in occluded areas} when training with more views, naturally providing a better training signal for occluded areas.
To qualitatively visualize these results we sample the camera frustum in horizontal slices from the center of the image downwards and aggregate the density in \cref{fig:profile}.
This shows the layout of the scene, similar to the birds-eye perspective but for density.
In the Full variant, the strong signal lets our model learn sharp object boundaries, as can be seen for several cars in the examples.
For depth prediction, all objects cast occupancy shadows along the viewing direction.
PixelNeRF predicts a volumetric representation with free space in occluded regions.
However, the results are noisy and the geometry is inaccurate.
MINE \cite{li2021mine} also specializes in predicting a volumetric representation from a single image. 
However, it does not produce meaningful density prediction behind objects.
Instead, similar to depth prediction, all objects cast occupancy shadows along the viewing direction.

\subsection{Depth Prediction}

\begin{table}
\centering
\footnotesize
\begin{tabular}{lccccc}
\toprule
\scriptsize\textit{Model} & \scriptsize\hspace{-.2cm}Volum.\ \hspace{-.2cm} & \scriptsize Split & \scriptsize Abs Rel $\downarrow$ & \scriptsize RMSE $\downarrow$ & \scriptsize\hspace{-.2cm}$\alpha < 1.25 \uparrow$\hspace{-.2cm} \\
\midrule
PixelNeRF \cite{yu2021pixelnerf} & \cmark & \hspace{-.2cm}\multirow{7}{*}{Eigen \cite{eigen2014depth}}\hspace{-.2cm} & 0.130 & 5.134 & 0.845 \\
EPC++ \cite{luo2019every}        & \xmark &  & 0.128 & 5.585 & 0.831 \\
MonoDepth2 \cite{godard2019digging}\hspace{-.2cm} & \xmark &  & 0.106 & 4.750 & 0.874 \\
PackNet \cite{guizilini20203d}   & \xmark &  & 0.111 & 4.601 & 0.878 \\
DepthHint \cite{watson2019self}  & \xmark &  & 0.105 & 4.627 & 0.875 \\
FeatDepth \cite{shu2020feature}  & \xmark &  & \underline{0.099} & 4.427 & \underline{0.889} \\
DevNet  \cite{zhou2022devnet}    & (\cmark) &  & \textbf{0.095} & \textbf{4.365} & \textbf{0.895} \\
\midrule
\textbf{Ours} & \cmark &  & 0.102 & \underline{4.407} & 0.882 \\ 
\midrule 
\midrule
MINE \cite{li2021mine} & \cmark & \hspace{-.2cm}\multirow{2}{*}{Tuls.\ \cite{tulsiani2018layer}}\hspace{-.2cm} & 0.137 & 6.592 & 0.839 \\
\textbf{Ours} & \cmark &  & \textbf{0.132} & \textbf{6.104} & \textbf{0.873} \\
\bottomrule
\end{tabular}
\caption{
\textbf{Depth Prediction on KITTI.} 
While our goal is full volumetric scene understanding, we compare to state-of-the-art self-supervised depth estimation method. 
Our approach achieves competitive performance while clearly improving over other volumetric approaches like PixelNeRF \cite{yu2021pixelnerf} and MINE \cite{li2021mine}.
DevNet \cite{zhou2022devnet} performs better, but does not show any results of their volume. 
\label{tab:depth_prediction}}
\vspace{-.2cm}
\end{table}
\begin{figure}
    \centering
    \includegraphics[width=.8\linewidth]{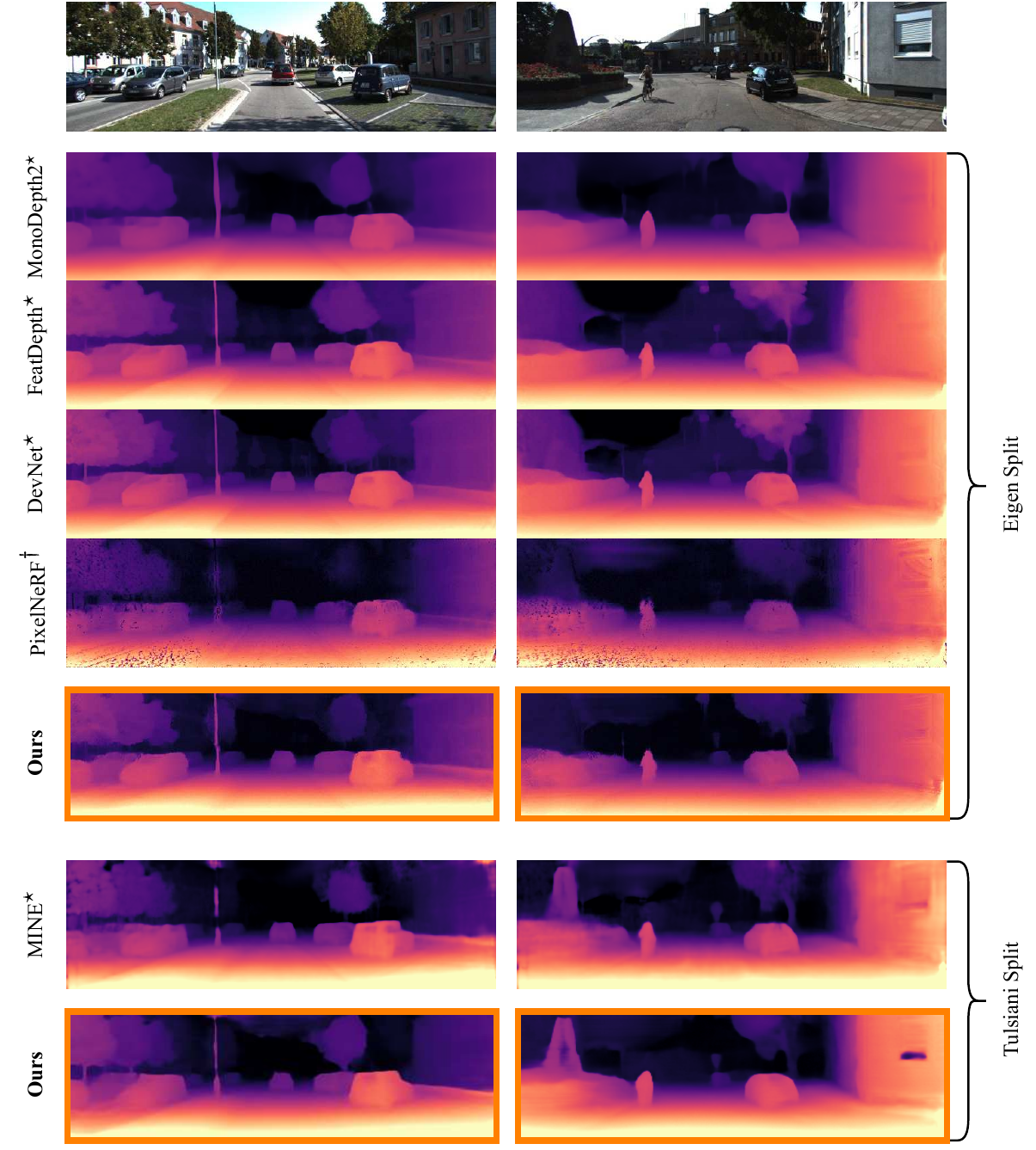}
    \caption{\textbf{Depth Prediction on KITTI.} Expected ray termination depth compared with depth prediction results of other state-of-the-art methods \cite{godard2019digging, shu2020feature, zhou2022devnet, li2021mine, yu2021pixelnerf} on both the Eigen \cite{eigen2014depth} and \cite{tulsiani2018layer} split. Our predictions are very detailed and sharp, and capture the structure of the scene, even when trained on a smaller split such as Tulsiani. Visualizations for DevNet and FeatDepth are taken from \cite{zhou2022devnet}. \textbf{Legend}: ref.\ \cref{fig:profile}.}
    \label{fig:depth_prediction}
    \vspace{-.2cm}
\end{figure}
While our method does not predict depth maps directly, they can be synthesized as a side product from our representation through the expected ray termination depth $\hat{d}$.
To demonstrate that our predicted representation achieves high accuracy, we train our model on KITTI sequences 
and compare to both self-supervised depth prediction methods and volume reconstruction methods.

As can be seen in \cref{tab:depth_prediction} and \cref{fig:depth_prediction}, our method performs on par with the current state-of-the-art methods for self-supervised depth prediction.
Our synthesized depth maps capture finer details and contain fewer artifacts, as often seen with depth maps obtained from neural radiance field-based methods, like PixelNeRF \cite{yu2021pixelnerf} and MINE \cite{li2021mine}.
Overall, we achieve competitive performance, even though depth prediction is not the main objective of our approach.

\subsection{Novel View Synthesis from a Single Image}
\begin{table*}[]
\centering
\footnotesize
\begin{tabular}{lccc|ccc|ccc}
\toprule
\multicolumn{1}{c}{} & \multicolumn{3}{c}{Configuration} & \multicolumn{3}{c}{Occupancy Estimation} & \multicolumn{3}{c}{Depth Prediction} \\
\midrule
\textit{Method} & Features & MLP & Predicts & $\text{O}_\text{acc} \uparrow$ & $\text{IE}_\text{acc} \uparrow$ & $\text{IE}_\text{rec} \uparrow$ & Abs Rel $\downarrow$ & RMSE $\downarrow$ & \hspace{-.2cm}$\alpha < 1.25 \uparrow$\hspace{-.2cm}\\ %
\midrule
\multirow{5}{*}{Baselines} & Enc & Big & $\sigma + c$ & 0.92 & 0.63 & \underline{0.41} & 0.130 & 5.134 & 0.845 \\
 & E+D & Big & $\sigma + c$ & 0.93 & 0.62 & \textbf{0.43} & 0.149 & 5.441 & 0.800 \\
 & Enc & Small & $\sigma + c$ & 0.92 & \underline{0.69} & 0.31 & 0.112 & 4.897 & 0.860 \\
 & E+D & Small & $\sigma + c$ & 0.93 & \underline{0.69} & 0.15 & 0.109 & 4.758 & 0.864 \\
 & Enc & Small & $\sigma$ & \textbf{0.94} & \textbf{0.77} & 0.39& \underline{0.105} & \underline{4.590} & \underline{0.872} \\
\textbf{Ours} & \textbf{E+D} & \textbf{Small} & \textbf{$\sigma$} & \textbf{0.94} & \textbf{0.77} & \textbf{0.43} & \textbf{0.102} & \textbf{4.407} & \textbf{0.882} \\
\midrule
\textbf{Ours} & \multicolumn{2}{l}{Keep invalid rays}& & 0.94 & 0.77 & 0.41 & 0.108 & 4.493 & 0.875  \\
\bottomrule
\end{tabular}
\caption{\textbf{Ablation Studies.} 
Evaluation of variants with different contributions (predicting only density $\sigma$ and sampling color, shifting capacity from the MLP to the feature extractor, discarding invalid rays) turned on / off.
Occupancy estimation results on KITTI-360 and depth prediction results on KITTI. 
The variant using only an encoder, big MLP, and color prediction corresponds exactly to the PixelNeRF \cite{yu2021pixelnerf} architecture, but with our training scheme.
\textbf{Legend}: \textit{Enc} Encoder, \textit{E+D} Encoder-Decoder, $\sigma$ density, $c$ color. \label{tab:ablations_combined}}
\vspace{-.4cm}
\end{table*}
\begin{table}[]
\centering
\footnotesize
\begin{tabular}{l|ccc|ccc}
\toprule
\multicolumn{1}{c}{} & \multicolumn{3}{c}{KITTI} & \multicolumn{3}{c}{RealEstate10K} \\
\midrule
\scriptsize \textit{Model} & \scriptsize \hspace{-.1cm}LPIPS $\downarrow$\hspace{-.1cm} & \scriptsize \hspace{-.1cm}SSIM $\uparrow$\hspace{-.1cm} & \scriptsize \hspace{-.2cm}PSNR $\uparrow$\hspace{-.2cm} & \scriptsize \hspace{-.1cm}LPIPS $\downarrow$\hspace{-.1cm} & \scriptsize \hspace{-.1cm}SSIM $\uparrow$\hspace{-.1cm} & \scriptsize \hspace{-.2cm}PSNR $\uparrow$\hspace{-.2cm} \\
\midrule
SynSin \cite{wiles2020synsin} & n/a & n/a & n/a & 1.180 & 0.740 & 22.3\\
Tulsiani \cite{tulsiani2018layer} & n/a & 0.572 & 16.5 & \underline{0.176} & \underline{0.785} & 23.5 \\
MPI \cite{tucker2020single} & n/a & 0.733 & 19.5 & n/a & n/a & n/a \\
MINE \cite{li2021mine} & \textbf{0.112} & \textbf{0.828} & \textbf{21.9} & \textbf{0.156} & \textbf{0.822} & \textbf{24.5} \\
PixelNeRF \cite{yu2021pixelnerf} & 0.175 & 0.761 & \underline{20.1} & n/a & n/a & n/a \\
\midrule
\textbf{Ours} & \underline{0.144} & \underline{0.764} & \underline{20.1} & 0.194 & 0.755 & \underline{24.0} \\
\bottomrule
\end{tabular}
\caption{\textbf{Novel View Synthesis.} We test the NVS ability on KITTI (Tulsiani split \cite{tulsiani2018layer}) and RealEstate10K (MINE split \cite{li2021mine}, target frame randomly sampled within 30 frames). Even though our method does not predict color, we still achieve strong results.\label{tab:nvs_combined}}
\vspace{-.4cm}
\end{table}

As we obtain a volumetric representation of a scene from a single image, we are able to synthesize images from novel viewpoints by sampling color from the input image.
Thus, we also evaluate novel view synthesis from a single image.
To demonstrate the variability of our approach, we train two models, one on RealEstate10K \cite{zhou2018stereo}, and one on the KITTI (Tulsiani split \cite{tulsiani2018layer}).
As \cref{tab:nvs_combined} shows, our method achieves strong performance on both datasets, despite the fact, that we only predict geometry and obtain color by sampling the input image.
Our results are comparable with many recent methods, that were specifically designed for this task, and of which some even use sparse depth supervision during training for RealEstate10K (MPI, MINE).
MINE \cite{li2021mine} achieves slightly better accuracy.
This can be attributed to them being able to predict color and thereby circumventing issues arising from imperfect geometry.

\subsection{Ablation Studies}

Our architectural design choices are critically important for the strong performance of our method.
To quantify the impact of the different contributions, we conduct ablation studies based on occupancy estimation on KITTI-360 and depth prediction on KITTI.
PixelNeRF \cite{yu2021pixelnerf} can be seen as a basis, which we modify step-by-step to reach our proposed model.
Namely, we 1.\ shift capacity from the MLP to the feature extractor and 2.\ introduce color sampling as an alternative to predicting the color alongside density.

As \cref{tab:ablations_combined} shows, reducing the MLP capacity and using a more powerful encoder-decoder rather than encoder as a feature extractor allows the model to learn significantly more precise overall geometry.
We conjecture that a powerful feature extractor is more suited to generalize to unseen scenes based on a single input image than a high-capacity MLP.
The feature extractor outputs a geometry representation (\ie the feature map) of the full scene in a single forward pass.
During training, it receives gradient information from all points sampled in the camera frustum, conditioned on the input image.
Thus, potential noise from small visual details gets averaged out.
On the other hand, the MLP outputs density based on the coordinates and is conditioned on a local feature.
The coordinates and feature are different for every sampled point, rather than per scene.
Consequently, noise will affect the MLP training significantly more. 

Introducing the sampling of color from the input frames further boosts accuracy, especially for occupancy estimation in occluded areas.
We hypothesize that only predicting density simplifies the training task significantly.
Crucially, the network does not have to hallucinate colors in occluded regions.
Additionally, color sampling enforces strict multi-view consistency.
The network cannot compensate for imperfect geometry by predicting the correct color.

Finally, the results show that our policy of discarding invalid rays during training improves accuracy by reducing noise in the training signal.
This mainly affects the border regions of the frustum.

\section{Conclusion}

In this paper, we introduced a new approach to learning to estimate the 3D geometric structure of a scene from a single image.
Our method predicts a continuous density field, which can be evaluated at any point in the camera frustum.
The key contributions in our paper are 1.\ color sampling, 2.\ architecture improvements, and 3.\ a new self-supervised loss formulation.
This enables us to train a network on large in-the-wild datasets with challenging scenes, such as KITTI, KITTI-360, and RealEstate10K.
We show that our method is able to capture geometry in occluded areas.
We evaluate depth maps synthesized from the predicted representation achieving comparable results to state-of-the-art methods. %
Despite only predicting geometry, our model even achieves high accuracy for novel view synthesis from a single image. %
Finally, we justify all of our design choices through detailed ablation studies.

\footnotesize{
\paragraph{Acknowledgements.}
This work was supported by the ERC Advanced Grant SIMULACRON, by the Munich Center for Machine Learning and by the EPSRC Programme Grant VisualAI EP/T028572/1.
C.\ R.\ is supported by VisualAI EP/T028572/1 and ERC-UNION-CoG-101001212.}

{\small
\bibliographystyle{ieee_fullname}
\bibliography{bibliography}

\begin{thebibliography}{10}\itemsep=-1pt

\bibitem{aich2021bidirectional}
Shubhra Aich, Jean Marie~Uwabeza Vianney, Md~Amirul Islam, and Mannat
  Kaur~Bingbing Liu.
\newblock Bidirectional attention network for monocular depth estimation.
\newblock In {\em 2021 IEEE International Conference on Robotics and Automation
  (ICRA)}, pages 11746--11752. IEEE, 2021.

\bibitem{barron2021mip}
Jonathan~T Barron, Ben Mildenhall, Matthew Tancik, Peter Hedman, Ricardo
  Martin-Brualla, and Pratul~P Srinivasan.
\newblock Mip-nerf: A multiscale representation for anti-aliasing neural
  radiance fields.
\newblock In {\em Proceedings of the IEEE/CVF International Conference on
  Computer Vision}, pages 5855--5864, 2021.

\bibitem{barron2022mip}
Jonathan~T Barron, Ben Mildenhall, Dor Verbin, Pratul~P Srinivasan, and Peter
  Hedman.
\newblock Mip-nerf 360: Unbounded anti-aliased neural radiance fields.
\newblock In {\em Proceedings of the IEEE/CVF Conference on Computer Vision and
  Pattern Recognition}, pages 5470--5479, 2022.

\bibitem{campos2021orb}
Carlos Campos, Richard Elvira, Juan J~G{\'o}mez Rodr{\'\i}guez, Jos{\'e}~MM
  Montiel, and Juan~D Tard{\'o}s.
\newblock Orb-slam3: An accurate open-source library for visual,
  visual--inertial, and multimap slam.
\newblock {\em IEEE Transactions on Robotics}, 37(6):1874--1890, 2021.

\bibitem{cao2022scenerf}
Anh-Quan Cao and Raoul de Charette.
\newblock Scenerf: Self-supervised monocular 3d scene reconstruction with
  radiance fields.
\newblock arxiv, 2022.

\bibitem{chen2022tensorf}
Anpei Chen, Zexiang Xu, Andreas Geiger, Jingyi Yu, and Hao Su.
\newblock Tensorf: Tensorial radiance fields.
\newblock {\em arXiv preprint arXiv:2203.09517}, 2022.

\bibitem{deng2022depth}
Kangle Deng, Andrew Liu, Jun-Yan Zhu, and Deva Ramanan.
\newblock Depth-supervised nerf: Fewer views and faster training for free.
\newblock In {\em Proceedings of the IEEE/CVF Conference on Computer Vision and
  Pattern Recognition}, pages 12882--12891, 2022.

\bibitem{dhamo2019object}
Helisa Dhamo, Nassir Navab, and Federico Tombari.
\newblock Object-driven multi-layer scene decomposition from a single image.
\newblock In {\em Proceedings of the IEEE/CVF International Conference on
  Computer Vision}, pages 5369--5378, 2019.

\bibitem{dhamo2019peeking}
Helisa Dhamo, Keisuke Tateno, Iro Laina, Nassir Navab, and Federico Tombari.
\newblock Peeking behind objects: Layered depth prediction from a single image.
\newblock {\em Pattern Recognition Letters}, 125:333--340, 2019.

\bibitem{eigen2014depth}
David Eigen, Christian Puhrsch, and Rob Fergus.
\newblock Depth map prediction from a single image using a multi-scale deep
  network.
\newblock {\em Advances in neural information processing systems}, 27, 2014.

\bibitem{fu2018deep}
Huan Fu, Mingming Gong, Chaohui Wang, Kayhan Batmanghelich, and Dacheng Tao.
\newblock Deep ordinal regression network for monocular depth estimation.
\newblock In {\em Proceedings of the IEEE conference on computer vision and
  pattern recognition}, pages 2002--2011, 2018.

\bibitem{geiger2013vision}
Andreas Geiger, Philip Lenz, Christoph Stiller, and Raquel Urtasun.
\newblock Vision meets robotics: The kitti dataset.
\newblock {\em The International Journal of Robotics Research},
  32(11):1231--1237, 2013.

\bibitem{godard2017unsupervised}
Cl{\'e}ment Godard, Oisin Mac~Aodha, and Gabriel~J Brostow.
\newblock Unsupervised monocular depth estimation with left-right consistency.
\newblock In {\em Proceedings of the IEEE conference on computer vision and
  pattern recognition}, pages 270--279, 2017.

\bibitem{godard2019digging}
Cl{\'e}ment Godard, Oisin Mac~Aodha, Michael Firman, and Gabriel~J Brostow.
\newblock Digging into self-supervised monocular depth estimation.
\newblock In {\em Proceedings of the IEEE/CVF International Conference on
  Computer Vision}, pages 3828--3838, 2019.

\bibitem{gonzalezbello2020forget}
Juan~Luis GonzalezBello and Munchurl Kim.
\newblock Forget about the lidar: Self-supervised depth estimators with med
  probability volumes.
\newblock {\em Advances in Neural Information Processing Systems},
  33:12626--12637, 2020.

\bibitem{guizilini20203d}
Vitor Guizilini, Rares Ambrus, Sudeep Pillai, Allan Raventos, and Adrien
  Gaidon.
\newblock 3d packing for self-supervised monocular depth estimation.
\newblock In {\em Proceedings of the IEEE/CVF Conference on Computer Vision and
  Pattern Recognition}, pages 2485--2494, 2020.

\bibitem{he2016deep}
Kaiming He, Xiangyu Zhang, Shaoqing Ren, and Jian Sun.
\newblock Deep residual learning for image recognition.
\newblock In {\em Proceedings of the IEEE conference on computer vision and
  pattern recognition}, pages 770--778, 2016.

\bibitem{jain2021putting}
Ajay Jain, Matthew Tancik, and Pieter Abbeel.
\newblock Putting nerf on a diet: Semantically consistent few-shot view
  synthesis.
\newblock In {\em Proceedings of the IEEE/CVF International Conference on
  Computer Vision}, pages 5885--5894, 2021.

\bibitem{kajiya1984ray}
James~T Kajiya and Brian~P Von~Herzen.
\newblock Ray tracing volume densities.
\newblock {\em ACM SIGGRAPH computer graphics}, 18(3):165--174, 1984.

\bibitem{kim2022infonerf}
Mijeong Kim, Seonguk Seo, and Bohyung Han.
\newblock Infonerf: Ray entropy minimization for few-shot neural volume
  rendering.
\newblock In {\em Proceedings of the IEEE/CVF Conference on Computer Vision and
  Pattern Recognition}, pages 12912--12921, 2022.

\bibitem{kuznietsov2017semi}
Yevhen Kuznietsov, Jorg Stuckler, and Bastian Leibe.
\newblock Semi-supervised deep learning for monocular depth map prediction.
\newblock In {\em Proceedings of the IEEE conference on computer vision and
  pattern recognition}, pages 6647--6655, 2017.

\bibitem{lee2021patch}
Sihaeng Lee, Janghyeon Lee, Byungju Kim, Eojindl Yi, and Junmo Kim.
\newblock Patch-wise attention network for monocular depth estimation.
\newblock In {\em Proceedings of the AAAI Conference on Artificial
  Intelligence}, volume~35, pages 1873--1881, 2021.

\bibitem{li2021mine}
Jiaxin Li, Zijian Feng, Qi She, Henghui Ding, Changhu Wang, and Gim~Hee Lee.
\newblock Mine: Towards continuous depth mpi with nerf for novel view
  synthesis.
\newblock In {\em Proceedings of the IEEE/CVF International Conference on
  Computer Vision}, pages 12578--12588, 2021.

\bibitem{li2022depthformer}
Zhenyu Li, Zehui Chen, Xianming Liu, and Junjun Jiang.
\newblock Depthformer: Exploiting long-range correlation and local information
  for accurate monocular depth estimation.
\newblock {\em arXiv preprint arXiv:2203.14211}, 2022.

\bibitem{li2022binsformer}
Zhenyu Li, Xuyang Wang, Xianming Liu, and Junjun Jiang.
\newblock Binsformer: Revisiting adaptive bins for monocular depth estimation.
\newblock {\em arXiv preprint arXiv:2204.00987}, 2022.

\bibitem{liao2022kitti}
Yiyi Liao, Jun Xie, and Andreas Geiger.
\newblock Kitti-360: A novel dataset and benchmarks for urban scene
  understanding in 2d and 3d.
\newblock {\em IEEE Transactions on Pattern Analysis and Machine Intelligence},
  2022.

\bibitem{liu2015learning}
Fayao Liu, Chunhua Shen, Guosheng Lin, and Ian Reid.
\newblock Learning depth from single monocular images using deep convolutional
  neural fields.
\newblock {\em IEEE transactions on pattern analysis and machine intelligence},
  38(10):2024--2039, 2015.

\bibitem{liu2020neural}
Lingjie Liu, Jiatao Gu, Kyaw Zaw~Lin, Tat-Seng Chua, and Christian Theobalt.
\newblock Neural sparse voxel fields.
\newblock {\em Advances in Neural Information Processing Systems},
  33:15651--15663, 2020.

\bibitem{luo2019every}
Chenxu Luo, Zhenheng Yang, Peng Wang, Yang Wang, Wei Xu, Ram Nevatia, and Alan
  Yuille.
\newblock Every pixel counts++: Joint learning of geometry and motion with 3d
  holistic understanding.
\newblock {\em IEEE transactions on pattern analysis and machine intelligence},
  42(10):2624--2641, 2019.

\bibitem{max1995optical}
Nelson Max.
\newblock Optical models for direct volume rendering.
\newblock {\em IEEE Transactions on Visualization and Computer Graphics},
  1(2):99--108, 1995.

\bibitem{mescheder2019occupancy}
Lars Mescheder, Michael Oechsle, Michael Niemeyer, Sebastian Nowozin, and
  Andreas Geiger.
\newblock Occupancy networks: Learning 3d reconstruction in function space.
\newblock In {\em Proceedings of the IEEE/CVF conference on computer vision and
  pattern recognition}, pages 4460--4470, 2019.

\bibitem{mildenhall2021nerf}
Ben Mildenhall, Pratul~P Srinivasan, Matthew Tancik, Jonathan~T Barron, Ravi
  Ramamoorthi, and Ren Ng.
\newblock Nerf: Representing scenes as neural radiance fields for view
  synthesis.
\newblock {\em Communications of the ACM}, 65(1):99--106, 2021.

\bibitem{muller2022autorf}
Norman M{\"u}ller, Andrea Simonelli, Lorenzo Porzi, Samuel~Rota Bul{\`o},
  Matthias Nie{\ss}ner, and Peter Kontschieder.
\newblock Autorf: Learning 3d object radiance fields from single view
  observations.
\newblock In {\em Proceedings of the IEEE/CVF Conference on Computer Vision and
  Pattern Recognition}, pages 3971--3980, 2022.

\bibitem{muller2022instant}
Thomas M{\"u}ller, Alex Evans, Christoph Schied, and Alexander Keller.
\newblock Instant neural graphics primitives with a multiresolution hash
  encoding.
\newblock {\em arXiv preprint arXiv:2201.05989}, 2022.

\bibitem{niemeyer2022regnerf}
Michael Niemeyer, Jonathan~T Barron, Ben Mildenhall, Mehdi~SM Sajjadi, Andreas
  Geiger, and Noha Radwan.
\newblock Regnerf: Regularizing neural radiance fields for view synthesis from
  sparse inputs.
\newblock In {\em Proceedings of the IEEE/CVF Conference on Computer Vision and
  Pattern Recognition}, pages 5480--5490, 2022.

\bibitem{Paszke_PyTorch_An_Imperative_2019}
Adam Paszke, Sam Gross, Francisco Massa, Adam Lerer, James Bradbury, Gregory
  Chanan, Trevor Killeen, Zeming Lin, Natalia Gimelshein, Luca Antiga, Alban
  Desmaison, Andreas Kopf, Edward Yang, Zachary DeVito, Martin Raison, Alykhan
  Tejani, Sasank Chilamkurthy, Benoit Steiner, Lu Fang, Junjie Bai, and Soumith
  Chintala.
\newblock {PyTorch: An Imperative Style, High-Performance Deep Learning
  Library}.
\newblock In {\em Advances in Neural Information Processing Systems 32}, pages
  8024--8035. Curran Associates, Inc., 2019.

\bibitem{peng2020convolutional}
Songyou Peng, Michael Niemeyer, Lars Mescheder, Marc Pollefeys, and Andreas
  Geiger.
\newblock Convolutional occupancy networks.
\newblock In {\em European Conference on Computer Vision}, pages 523--540.
  Springer, 2020.

\bibitem{roessle2022dense}
Barbara Roessle, Jonathan~T Barron, Ben Mildenhall, Pratul~P Srinivasan, and
  Matthias Nie{\ss}ner.
\newblock Dense depth priors for neural radiance fields from sparse input
  views.
\newblock In {\em Proceedings of the IEEE/CVF Conference on Computer Vision and
  Pattern Recognition}, pages 12892--12901, 2022.

\bibitem{sajjadi2022scene}
Mehdi~SM Sajjadi, Henning Meyer, Etienne Pot, Urs Bergmann, Klaus Greff, Noha
  Radwan, Suhani Vora, Mario Lu{\v{c}}i{\'c}, Daniel Duckworth, Alexey
  Dosovitskiy, et~al.
\newblock Scene representation transformer: Geometry-free novel view synthesis
  through set-latent scene representations.
\newblock In {\em Proceedings of the IEEE/CVF Conference on Computer Vision and
  Pattern Recognition}, pages 6229--6238, 2022.

\bibitem{schoenberger2016sfm}
Johannes~Lutz Sch\"{o}nberger and Jan-Michael Frahm.
\newblock Structure-from-motion revisited.
\newblock In {\em Conference on Computer Vision and Pattern Recognition
  (CVPR)}, 2016.

\bibitem{schoenberger2016mvs}
Johannes~Lutz Sch\"{o}nberger, Enliang Zheng, Marc Pollefeys, and Jan-Michael
  Frahm.
\newblock Pixelwise view selection for unstructured multi-view stereo.
\newblock In {\em European Conference on Computer Vision (ECCV)}, 2016.

\bibitem{shade1998layered}
Jonathan Shade, Steven Gortler, Li-wei He, and Richard Szeliski.
\newblock Layered depth images.
\newblock In {\em Proceedings of the 25th annual conference on Computer
  graphics and interactive techniques}, pages 231--242, 1998.

\bibitem{sharma2022seeing}
Prafull Sharma, Ayush Tewari, Yilun Du, Sergey Zakharov, Rares Ambrus, Adrien
  Gaidon, William~T Freeman, Fredo Durand, Joshua~B Tenenbaum, and Vincent
  Sitzmann.
\newblock Seeing 3d objects in a single image via self-supervised
  static-dynamic disentanglement.
\newblock {\em arXiv preprint arXiv:2207.11232}, 2022.

\bibitem{shu2020feature}
Chang Shu, Kun Yu, Zhixiang Duan, and Kuiyuan Yang.
\newblock Feature-metric loss for self-supervised learning of depth and
  egomotion.
\newblock In {\em European Conference on Computer Vision}, pages 572--588.
  Springer, 2020.

\bibitem{silberman2012indoor}
Nathan Silberman, Derek Hoiem, Pushmeet Kohli, and Rob Fergus.
\newblock Indoor segmentation and support inference from rgbd images.
\newblock In {\em European conference on computer vision}, pages 746--760.
  Springer, 2012.

\bibitem{srinivasan2019pushing}
Pratul~P Srinivasan, Richard Tucker, Jonathan~T Barron, Ravi Ramamoorthi, Ren
  Ng, and Noah Snavely.
\newblock Pushing the boundaries of view extrapolation with multiplane images.
\newblock In {\em Proceedings of the IEEE/CVF Conference on Computer Vision and
  Pattern Recognition}, pages 175--184, 2019.

\bibitem{takikawa2021neural}
Towaki Takikawa, Joey Litalien, Kangxue Yin, Karsten Kreis, Charles Loop, Derek
  Nowrouzezahrai, Alec Jacobson, Morgan McGuire, and Sanja Fidler.
\newblock Neural geometric level of detail: Real-time rendering with implicit
  3d shapes.
\newblock In {\em Proceedings of the IEEE/CVF Conference on Computer Vision and
  Pattern Recognition}, pages 11358--11367, 2021.

\bibitem{tucker2020single}
Richard Tucker and Noah Snavely.
\newblock Single-view view synthesis with multiplane images.
\newblock In {\em Proceedings of the IEEE/CVF Conference on Computer Vision and
  Pattern Recognition}, pages 551--560, 2020.

\bibitem{tulsiani2018layer}
Shubham Tulsiani, Richard Tucker, and Noah Snavely.
\newblock Layer-structured 3d scene inference via view synthesis.
\newblock In {\em Proceedings of the European Conference on Computer Vision
  (ECCV)}, pages 302--317, 2018.

\bibitem{wang2004image}
Zhou Wang, Alan~C Bovik, Hamid~R Sheikh, and Eero~P Simoncelli.
\newblock Image quality assessment: from error visibility to structural
  similarity.
\newblock {\em IEEE transactions on image processing}, 13(4):600--612, 2004.

\bibitem{watson2019self}
Jamie Watson, Michael Firman, Gabriel~J Brostow, and Daniyar Turmukhambetov.
\newblock Self-supervised monocular depth hints.
\newblock In {\em Proceedings of the IEEE/CVF International Conference on
  Computer Vision}, pages 2162--2171, 2019.

\bibitem{wiles2020synsin}
Olivia Wiles, Georgia Gkioxari, Richard Szeliski, and Justin Johnson.
\newblock Synsin: End-to-end view synthesis from a single image.
\newblock In {\em Proceedings of the IEEE/CVF Conference on Computer Vision and
  Pattern Recognition}, pages 7467--7477, 2020.

\bibitem{wimbauer2022rendering}
Felix Wimbauer, Shangzhe Wu, and Christian Rupprecht.
\newblock De-rendering 3d objects in the wild.
\newblock In {\em Proceedings of the IEEE/CVF Conference on Computer Vision and
  Pattern Recognition}, pages 18490--18499, 2022.

\bibitem{wimbauer2021monorec}
Felix Wimbauer, Nan Yang, Lukas Von~Stumberg, Niclas Zeller, and Daniel
  Cremers.
\newblock Monorec: Semi-supervised dense reconstruction in dynamic environments
  from a single moving camera.
\newblock In {\em Proceedings of the IEEE/CVF Conference on Computer Vision and
  Pattern Recognition}, pages 6112--6122, 2021.

\bibitem{wu2020unsupervised}
Shangzhe Wu, Christian Rupprecht, and Andrea Vedaldi.
\newblock Unsupervised learning of probably symmetric deformable 3d objects
  from images in the wild.
\newblock In {\em Proceedings of the IEEE/CVF Conference on Computer Vision and
  Pattern Recognition}, pages 1--10, 2020.

\bibitem{yang2018deep}
Nan Yang, Rui Wang, Jorg Stuckler, and Daniel Cremers.
\newblock Deep virtual stereo odometry: Leveraging deep depth prediction for
  monocular direct sparse odometry.
\newblock In {\em Proceedings of the European Conference on Computer Vision
  (ECCV)}, pages 817--833, 2018.

\bibitem{yu2021pixelnerf}
Alex Yu, Vickie Ye, Matthew Tancik, and Angjoo Kanazawa.
\newblock pixelnerf: Neural radiance fields from one or few images.
\newblock In {\em Proceedings of the IEEE/CVF Conference on Computer Vision and
  Pattern Recognition}, pages 4578--4587, 2021.

\bibitem{yuan2022new}
Weihao Yuan, Xiaodong Gu, Zuozhuo Dai, Siyu Zhu, and Ping Tan.
\newblock New crfs: Neural window fully-connected crfs for monocular depth
  estimation.
\newblock {\em arXiv preprint arXiv:2203.01502}, 2022.

\bibitem{zhan2018unsupervised}
Huangying Zhan, Ravi Garg, Chamara~Saroj Weerasekera, Kejie Li, Harsh Agarwal,
  and Ian Reid.
\newblock Unsupervised learning of monocular depth estimation and visual
  odometry with deep feature reconstruction.
\newblock In {\em Proceedings of the IEEE conference on computer vision and
  pattern recognition}, pages 340--349, 2018.

\bibitem{zhou2022devnet}
Kaichen Zhou, Lanqing Hong, Changhao Chen, Hang Xu, Chaoqiang Ye, Qingyong Hu,
  and Zhenguo Li.
\newblock Devnet: Self-supervised monocular depth learning via density volume
  construction.
\newblock {\em arXiv preprint arXiv:2209.06351}, 2022.

\bibitem{zhou2017unsupervised}
Tinghui Zhou, Matthew Brown, Noah Snavely, and David~G Lowe.
\newblock Unsupervised learning of depth and ego-motion from video.
\newblock In {\em Proceedings of the IEEE conference on computer vision and
  pattern recognition}, pages 1851--1858, 2017.

\bibitem{zhou2018stereo}
Tinghui Zhou, Richard Tucker, John Flynn, Graham Fyffe, and Noah Snavely.
\newblock Stereo magnification: Learning view synthesis using multiplane
  images.
\newblock {\em arXiv preprint arXiv:1805.09817}, 2018.

\bibitem{zhou2016view}
Tinghui Zhou, Shubham Tulsiani, Weilun Sun, Jitendra Malik, and Alexei~A Efros.
\newblock View synthesis by appearance flow.
\newblock In {\em European conference on computer vision}, pages 286--301.
  Springer, 2016.

\end{thebibliography}
}

\clearpage
\appendix
\renewcommand{\thesection}{\Alph{section}}
\renewcommand{\thesubsection}{\Alph{section}.\arabic{subsection}}

\section{Ethics}

This research uses datasets (KITTI \cite{geiger2013vision}, KITTI-360 \cite{liao2022kitti}, and RealEstate10K \cite{zhou2018stereo}) to develop and benchmark computer vision models. 
The datasets are used in a manner compatible with their terms of usage. 
Some datasets the images can contain visible faces and other personal data collected without consent, however there is no processing of biometric information. 
Images are CC-BY or used in a manner compatible with the Data Analysis Permission.
We do not process biometric information. 
Please see \url{https://www.robots.ox.ac.uk/~vedaldi/research/union/ethics.html} for further information on ethics and data protection rights and mitigation.

\section{Limitations}
As the model makes predictions from a single frame, it can only rely on priors to predict visible \textit{and invisible} parts of the scene. 
Naturally, it should thus not be used in safety-critical applications, nor outside of a research setting. 

As the model is sampling colors instead of predicting them, it cannot predict plausible colors for unseen objects. Thus, in the setting of novel view synthesis, extreme camera pose changes tend to lead to visible artifacts in the images.

Similar to self-supervised depth prediction methods, our loss formulation relies on photometric consistency.
View-dependent effects are not modeled explicitly and could therefore introduce noise in the training process.

Further, our loss formulation relies on a static scene assumption and dynamic objects are not modeled explicitly.
While this has the potential to reduce accuracy, there are several reasons why it only has marginal effect in our case.
1.\ In most cases, we have stereo frames available, which give accurate training signals, even for moving objects. 2.\ The time difference between the different views is very small (usually in the order of $0.1$ seconds.
Therefore, even if an object is moving, the introduced noise is rather small.
Nonetheless, it would be an interesting direction for future work to investigate explicit modelling of dynamic objects in a loss formulation like ours.

\section{Additional Results}

In the following, we show additional results for novel view synthesis, capturing true 3D, and depth prediction. Please also see the video for additional qualitative results and explanations.
Figures can be found after the text of the supplementary material.

\subsection{Novel View Synthesis}

\begin{figure*}
    \centering
    \includegraphics[trim={0cm .3cm 0cm 0cm},width=\linewidth]{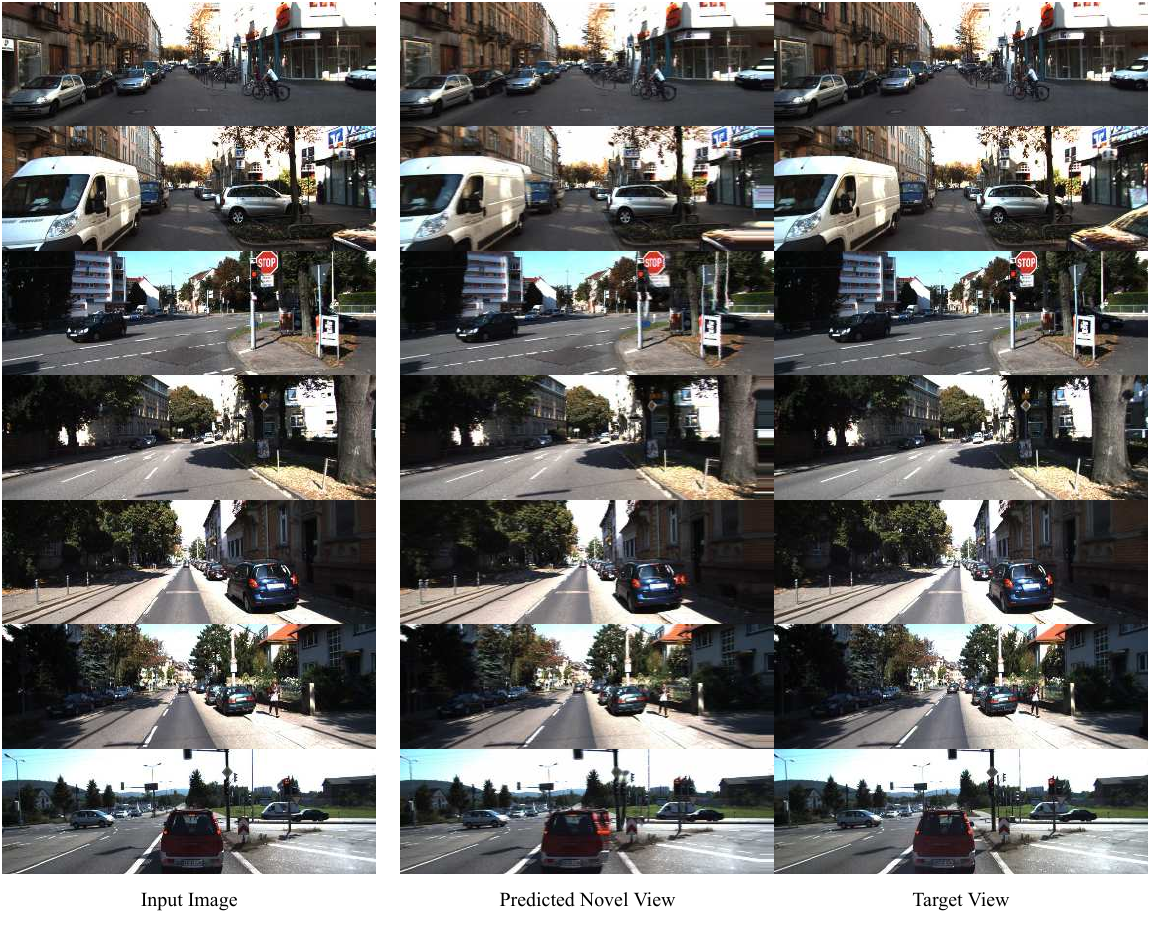}
    \caption{\textbf{Novel View Synthesis on KITTI.} Rendering the right stereo frame based on the density field predicted from the left stereo frame. Colors are also sampled from the same frame we make the prediction from. Areas of the image that are not occluded in both the input and target frame are reconstructed very accurately.}
    \label{fig:nvs_kitti}
\end{figure*}

\begin{figure*}
    \centering
    \includegraphics[trim={0cm .3cm 0cm 0cm},width=\linewidth]{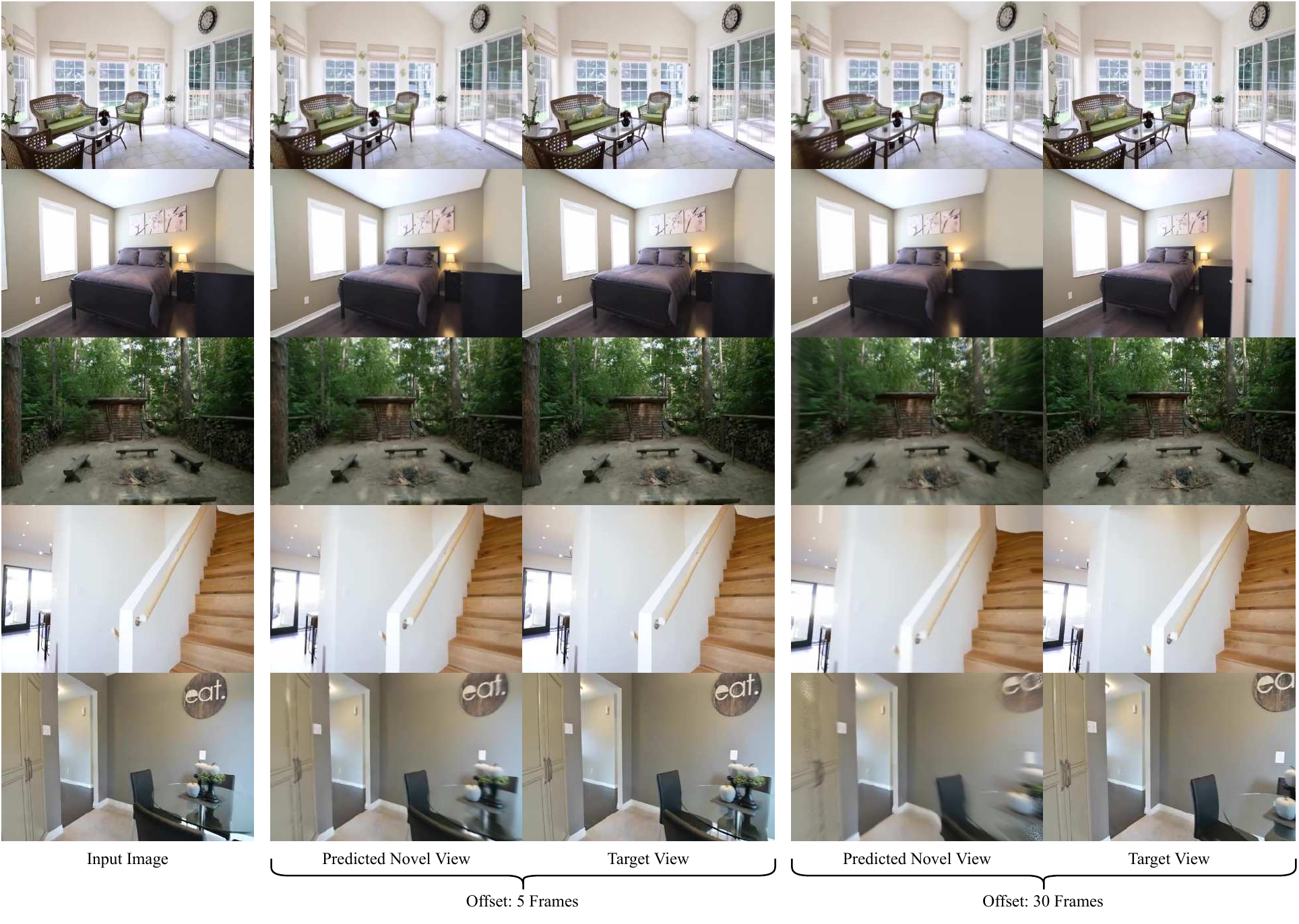}
    \caption{\textbf{Novel View Synthesis on RealEstate10K.} Rendering a later frame based on the density field predicted from the input frame. Colors are also sampled from the same frame we make the prediction from.
    }
    \label{fig:nvs_re10k}
\end{figure*}

\cref{fig:nvs_kitti} shows qualitative results from the Tulsiani \cite{tulsiani2018layer} test split for KITTI \cite{geiger2013vision}.
\cref{fig:nvs_re10k} shows qualitative results for the test set of RealEstate10K \cite{zhou2018stereo} proposed by MINE \cite{li2021mine}.

\subsection{Capturing True 3D}

\begin{figure*}
    \centering
    \includegraphics[trim={0cm .2cm 0cm 0cm},width=\linewidth]{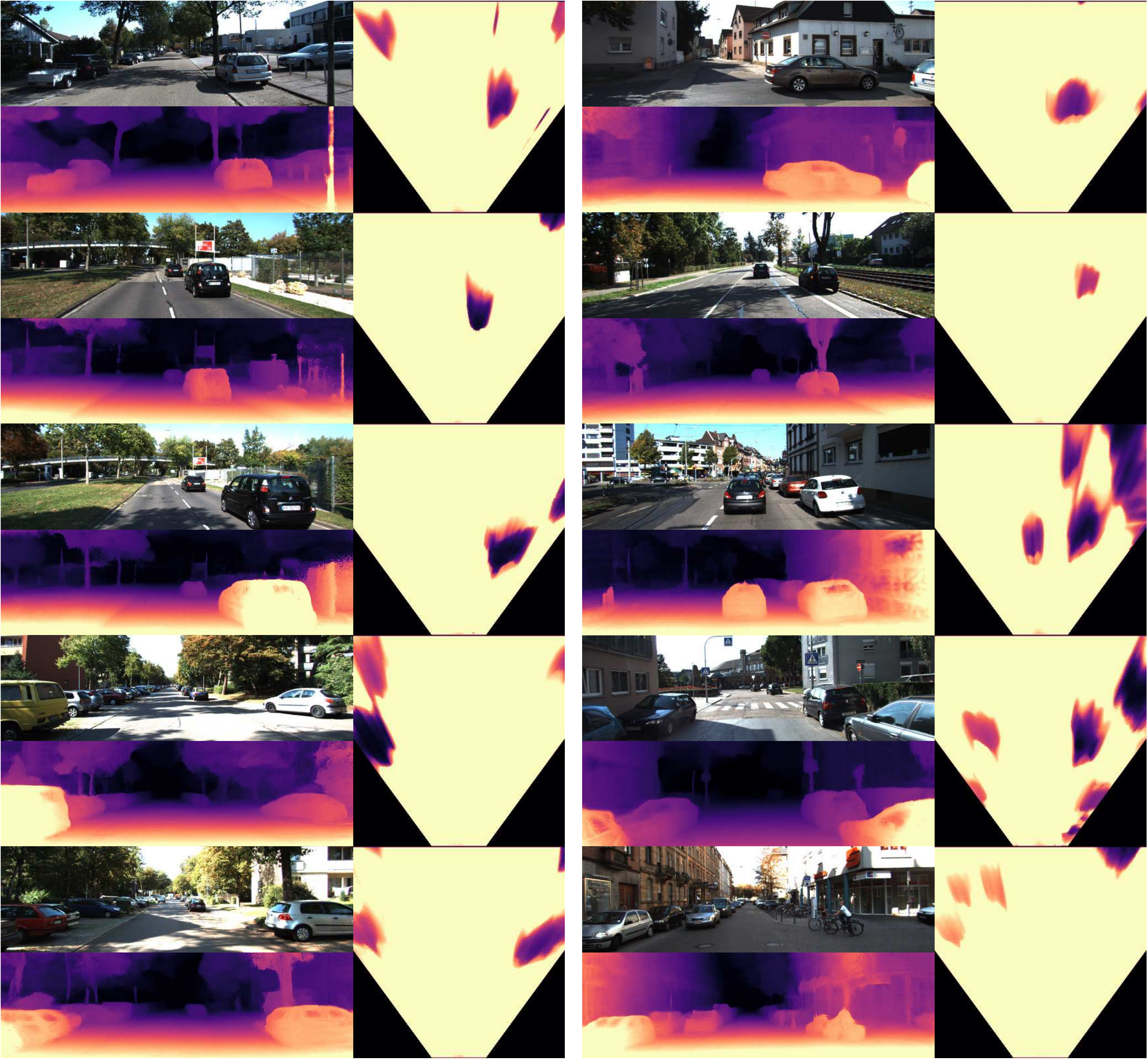}
    \caption{\textbf{Occupancy Estimation.} More qualitative top-down visualization of the occupancy map predicted by different methods. We show an area of $x=[-15m, 15m], z=[5m, 30m]$ and aggregate density from the $y$-coordinate of the camera $1m$ downward.}
    \label{fig:profile_supp}
\end{figure*}

\cref{fig:profile_supp} shows further visualizations of the predicted density field, in which you can clearly make out the different objects in the scene in the top-down view.

\subsection{Depth Prediction}

\begin{figure*}
    \centering
    \includegraphics[width=\linewidth]{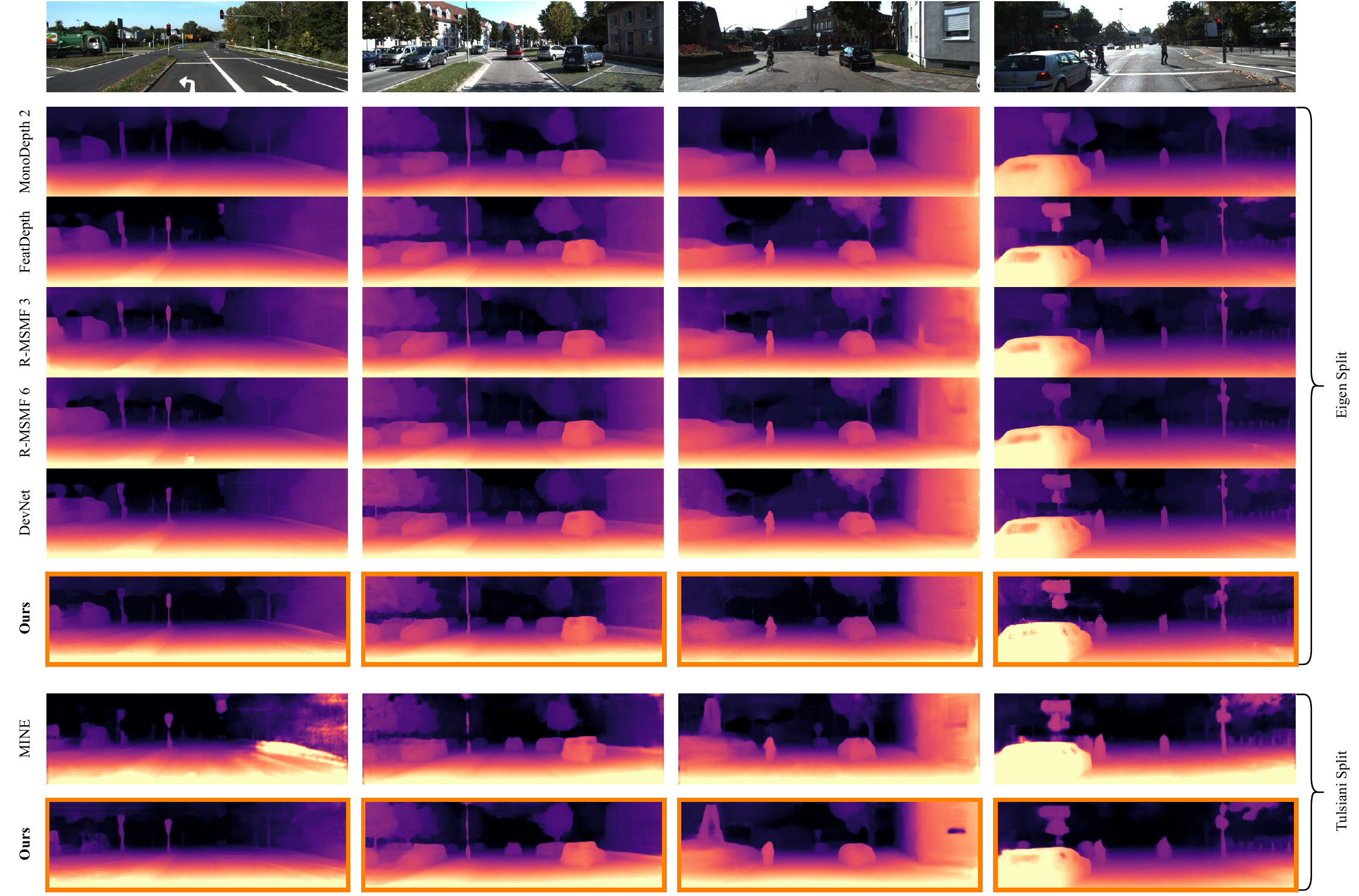}
    \caption{\textbf{Depth Prediction.} Additional visualizations of the expected ray termination depth compared with depth prediction results of other state-of-the-art methods \cite{godard2019digging, shu2020feature, zhou2022devnet, li2021mine} on both the Eigen \cite{eigen2014depth} and \cite{tulsiani2018layer} split. Visualizations for DevNet and FeatDepth are taken from \cite{zhou2022devnet}.}
    \label{fig:depth_prediction_supp}
\end{figure*}

\cref{fig:depth_prediction_supp} shows further results comparing our expected ray termination depth with results from other depth prediction methods.
\cref{tab:depth_prediction_full} reports additional metrics to the table in the main paper.

\section{Technical Details}

In the following, we discuss the exact implementation details, network configurations, and training setup, so that our results can be reproduced easily.
Further, we provide further details regarding the computation of the occupancy metrics.

\subsection{Implementation Details}

We base our implementation on the official code repository published by \cite{yu2021pixelnerf}.
Further, we are inspired by the repository of \cite{godard2019digging} regarding the implementation of the image reconstruction loss functions.

\paragraph{Networks.}
For encoder, we use a ResNet-50 \cite{he2016deep} backbone pretrained on ImageNet. 
We rely on the official weights provided by PyTorch \cite{Paszke_PyTorch_An_Imperative_2019} / Torchvision.
As decoder, we follow the architecture of MonoDepth2 \cite{godard2019digging} with a minor modification. 
Since we output feature maps with $C$ channels at the same resolution of the input, we do not reduce the features during upconvolutions below C to prevent information loss.
Concretely for every layer, we have an output channel dimension of $\hat{C}_\text{out} = \max(C, C_\text{out})$, where $C_\text{out}$ is the output channel dimension of the MonoDepth2 model.
We found $C=64$ to give best results.

For the decoding MLP, we use two fully-connected layers with hidden dimension $C$ (same as the feature dimension) and ReLU activation function.
We found that more layers do not improve the quality of the reconstruction.
Our hypothesis is that the decoding is a simple task that does not require a network with high capacity.

\paragraph{Rendering.}
To obtain a color / expected ray termination depth for a given ray, we sample $S$ points between $z_\text{near}$ and $z_\text{far}$.
As we deal with potentially unbounded scenes with many different scales, we use inverse depth to obtain the ranges for the different parts.
For every range, we uniformly draw one sample.
Let $d_i$ be the depth step for the $i$-th point ($i \in [0, S-1]$) and $r\sim U[0, 1]$ a random sample from the uniform distribution between $0$ and $1$.
\begin{equation}
    d_i = 1/\left(\frac{1 - s_i}{z_\text{near}} + \frac{s_i}{z_\text{far}}\right), \quad s_i = \frac{i + r}{S}
\end{equation}

We also experimented with coarse and fine sampling as used in many NeRF papers (\eg \cite{yu2021pixelnerf, mildenhall2021nerf}).
Here, after sampling the entire range of depths as above (coarse sampling), we perform importance sampling based on the returned weights and sampling around the expected ray termination depth (fine sampling).
Further, we also duplicate the MLP: one for coarse and one for fine sampling.
While the outputs of both networks are used for two seperate reconstructions with separate losses, only the fine reconstruction results are used for evaluation.
This technique is particularly helpful in NeRFs to increase the visual quality.
While the coarse MLP has to model the density and color distribution for a big range of coordinates, the fine MLP only has to learn the relevant area around surfaces.
In our experiments, we found that we do not get any benefit from adding fine sampling, both when using two seperate MLPs or one for coarse and fine sampling.
We suspect that our single MLP already has enough capacity to model the density distribution (we do not model color with the MLP) described by the feature at sufficient accuracy.

\paragraph{Positional Encoding.}
As described in the main paper, we pass $d_i$ and $\textbf{u}_\textbf{I}^\prime$ (pixel coordinate) values through a positional encoding function, before feeding them to the network along side the sampled feature $f_{\textbf{u}_\textbf{I}^\prime}$.
This positional encoding functions maps the input to $\sin$ and $\cos$ functions with different frequencies.
This is an established practice in methods where networks have to reason about the spatial location of points in 2D or 3D \cite{yu2021pixelnerf, mildenhall2021nerf}.
As we deal with real-world scale of scenes, we first normalize the depth to $[-1, 1]$.
This ensures that the data-range perfectly matches the used frequencies.
$\textbf{u}_\textbf{I}^\prime$ uses normalized pixel coordinates with $\textbf{u}_\textbf{I}^\prime \in [-1, 1]^2$ already.
For a vector, we compute the positional encoding per element as: 
\begin{align}
\begin{split}
    \gamma(x) = [&x, \sin(x \pi 2^0), \cos(x \pi 2^0), \sin(x \pi 2^1), \cos(x \pi 2^1),\\  
                 &\dots, \sin(x \pi 2^6), \cos(x \pi 2^6)]
\end{split}
\end{align}

\subsection{Training Configuration}

\begin{figure}
    \centering
    \includegraphics[trim={0cm .2cm 0cm 0cm},width=\linewidth]{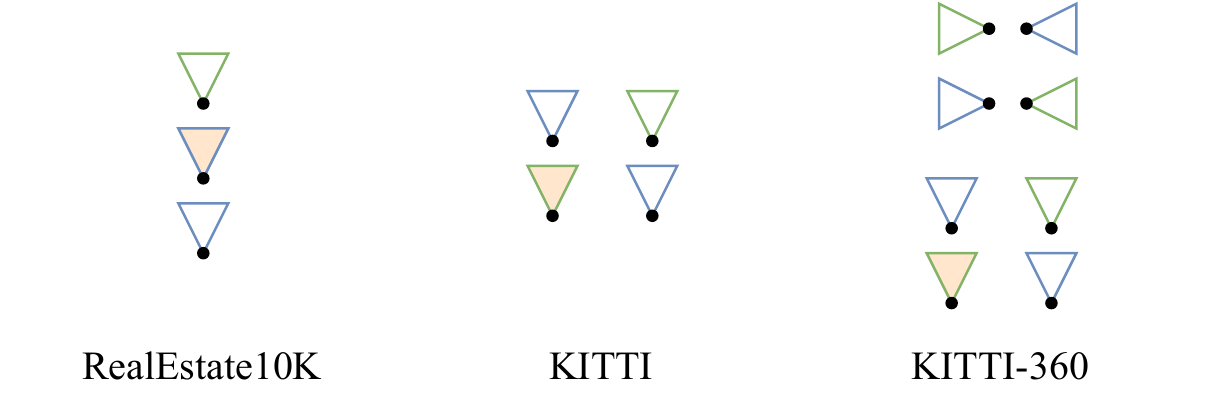}
    \caption{\textbf{Frame Arrangement per Sample.} 
    RealEstate10K only has monocular sequences. 
    KITTI and KITTI-360 provide stereo. 
    KITTI-360 also contains fisheye camera frames facing left and right. \textbf{Legend}: ref.\ Fig.\ 3.}
    \label{fig:data_setup}
\end{figure}

Through preliminary experiments, we found that the following training configuration yields the best results:

In all of our experiments, we use a batch size of 16.
In total, we sample 2048 rays for each item in a batch. 
These rays are grouped in $8 \times 8$ sized patches randomly sampled from any of the frames in $N_\text{loss}$
From this, the loss is computed.
Following \cite{godard2019digging}, we set $\lambda_\text{SSIM} = 0.85$, $\lambda_\text{L1} = 0.15$ and $\lambda_\text{EAS} = 0.001 * 2$. %
The default learning rate is $\lambda = 10^{-4}$ and we decrease it to $\lambda = 10^{-5}$ for the last 20\% of the training.
For all trainings, we use color augmentation (same parameters for all views of an item in the batch) and flip augmentation (randomly horizontally flip the image that gets fed into the encoder-decoder and then flip the resulting feature maps back to avoid changing the geometry of the scene).
\cref{tab:data_overview} shows an overview of the different datasets and the used split.
\cref{fig:data_setup} visualizes the frame arrangement and a possible partitioning into $N_\text{loss}$ and $N_\text{render}$ for the different datasets.
\begin{table}[]
\centering
\footnotesize
\begin{tabular}{lcccc}
\toprule
\textit{Dataset} & Split & \#Train & \#Val.\ & \#Test \\
\midrule
\multirow{2}{*}{KITTI \cite{geiger2013vision}} & Eigen \cite{eigen2014depth} & 39.810 & 4.424 & 697 \\
& Tulsiani \cite{tulsiani2018layer} & 11.987 & 1.243 & 1.079 \\
\midrule
KITTI-360 \cite{liao2022kitti} & Ours & 98.008 & 11.451 & 446 \\
\midrule
RealEstate10K \cite{zhou2018stereo} & MINE \cite{li2021mine} & 8.954.743 & 245 & 3270 \\
\bottomrule
\end{tabular}
\caption{\textbf{Dataset Overview.} Different datasets used in this work with information on data split. Our KITTI-360 split is a modified version of the split for the image segmentation task. \label{tab:data_overview}}
\vspace{-.1cm}
\end{table}

\paragraph{KITTI.}
We rely on poses computed from ORB-SLAM 3 \cite{campos2021orb}, which uses the given stereo cameras, intrinsics, and baseline length.
To be consistent with popular depth prediction methods, we perform all trainings and experiments at a resolution of $640 \times 192$ and rely on the Eigen split \cite{eigen2014depth}.
For evaluation, like in most other works, the cut off distance is set to $80$m.
The training runs for 50 epochs and we reduce the learning rate after $100.000$ iterations.
We report depth prediction results for our model which was trained with two timesteps (input + following) and stereo, \ie four frames in total.
For occupancy estimation, we also train a model with three timesteps (previous + input + following) and stereo.
Depth prediction results for this model are on par, but not better than the model trained with two timesteps.
A depth range of $z_\text{near} = 3$m and $z_\text{far} = 80$m proved to work best.

On the Tulsiani split \cite{tulsiani2018layer}, we use the same settings, except that we train for 150 epochs, as the split contains around $3\times$ fewer samples.

Training in both cases takes around four days.

\paragraph{KITTI-360.}
As the setting is very similar to KITTI, we use the same parameters.
Because the dataset is significantly larger, we only train for 25 epochs.
Training again takes around four days.
Additionally to the two stereo frames, we also have access to fisheye camera pointing left and right.
In order to be able to use them within our implementation, we resample them based on a virtual perspective camera with the same parameters as the forward-facing perspective cameras.
Note that the fisheye cameras seem to be mounted higher up than the perspective cameras.
Therefore, we rotate the virtual cameras $15^\circ$ downwards along the $x$-axis during the resampling process.
Further, fisheye and forward-facing cameras of the same timeframe have barely any overlapping visible areas.
Therefore, we offset fisheye cameras by 10 timesteps.
\begin{figure}
    \centering
    \includegraphics[width=\linewidth]{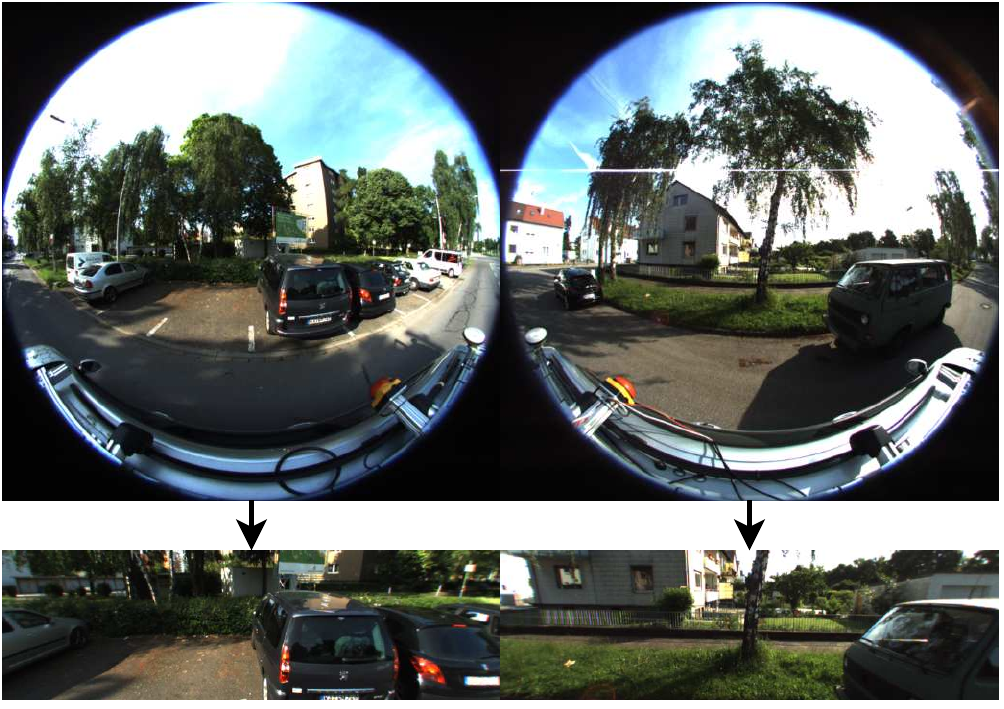}
    \caption{\textbf{Fisheye Resampling.} KITTI-360 provides frames from two fisheye cameras, one facing to the left and one facing to the right. We resample them based on a virtual perspective camera that has the same camera intrinsics as the perspective forward-facing cameras. We rotate the virtual camera $15^\circ$ downward to maximize the overlap of the frustums with the forward-facing cameras.}
    \label{fig:fisheye_resampling}
\end{figure}
\cref{fig:fisheye_resampling} shows examples from the fisheye cameras and the resampled images.

\paragraph{RealEstate10K}
As RealEstate10K contains magnitudes more images than KITTI or KITTI-360, approximately 8 mio.\ for training.
Therefore, we define the training length by the number of iterations, in this case 360k.
We follow \cite{li2021mine} and perform all experiments at a resolution of $384 \times 256$.
We train with three frames (previous + input + following) per item in the batch.
As the framerate is very high, we randomly draw an offset from the range $[1, 30]$ between the frames.
\cite{zhou2018stereo} states that all sequences are normalized to fit a depth range of $z_\text{near} = 1m$ and $z_\text{far} = 100m$ with inverse depth spacing.
We use the poses provided by the dataset.

\subsection{Occupancy Metric}

We rely on Lidar scans provided by KITTI-360 to build ground-truth occupancy and visibility maps, which we then use to evaluate our prediction quality.
Our evaluation protocol relies on 2D slices parallel to the ground between the street and the height of the car.
This allows to focus on the interesting regions of the scene, that also contain other objects like cars and pedestrians, and ignore areas that are not interesting, like the area below the street or the sky.

Consider now a single input frame, for which we would like to build our ground-truth occupancy and visibility.
As KITTI-360 is an autonomous driving dataset, the vehicle is generally moving forward at a steady pace.
Thus, consecutive Lidar scans captured a short time later measure different areas within the camera frustum.
Note that these Lidar measurements can reach areas that are occluded in the input image.
To determine whether a point is occupied, we check whether it is \textit{in front} of the measured surface for any of the Lidar scans.
Intuitively, every Lidar measurement ``carves out'' unoccupied areas in 3D space.

Let $L_i = \left\{\textbf{x}_j \in \mathrm{R}^3 | j \in [M] \right\}$ be the Lidar scan $i$ timesteps after the input frame with $M$ measurement points in the world coordinate system.
$L_0$ denotes the Lidar scan captured synchronously to the input frame.
Let $P_i$ denote the vehicle-to-world transformation (for ease of notation we assume that the Lidar scanner is centered at the vehicle pose and that the input frame has the same pose as the the $0$-th Lidar scan).

3D interpolation with sparse Lidar point clouds is difficult.
Therefore, we extract slices from the scan pointclouds and project them onto a 2D plane.
Additionally, we convert every point from Cartesian coordinates to polar coordinates, centered around the origin of the respective Lidar scan.
This makes the measurements much more dense and evaluation of whether a point is in front or behind a surface easier.

Let $y_\text{min}, y_\text{max}$ describe the min and max y-coordinate for our slice.
$\text{pol}_{xz}(\textbf{x}) \rightarrow (\alpha, d)$ denotes a function to convert a Cartesian coordinate to a polar coordinate after projecting it onto the xz-plane.
\begin{equation}
    S_i = \left\{\text{pol}_{xz}(\textbf{x}_j) | \textbf{x}_j \in L_i \wedge y_\text{min} \leq x_j^y \leq y_\text{max} \right\}
\end{equation}

For a given Lidar scan, we can now check whether a point $\textbf{x}$ is in front or behind the measured surface by transforming it into the scan's coordinate system, converting it to polar coordinates and then comparing the distance.
However, experiments showed that the Lidar scans in KITTI-360 can be fairly noisy, especially for objects like cars, that have translucent or reflective materials. 
Oftentimes, single outlier points are measured to be at a much bigger distance.
As we rely on the ``carving out'' idea, such points would carve out a lot of free space and lead to inaccurate occupancy maps.
To filter out these outliers, we split the 360 degree range of the polar coordinates into $b = 360$ equally sized bins and assign every measured point to the corresponding bin.
For every bin $S_i[\alpha]$, we then choose the minimal measured distance.

Using the $S_i$, we now define a function that allows to check whether a point is occupied or not.
Note that we can obtain the two closest bins for a given angle $\alpha$ through the floor and ceil functions.
\begin{equation}
\begin{split}
    \alpha, d &= \text{pol}_{xz}(P_i^{-1} \textbf{x}) \\
    \alpha_l, \alpha_r &= \left\lfloor \alpha \right\rfloor, \left\lceil \alpha \right\rceil \\
    \delta &= \frac{\alpha - \alpha_l}{\alpha_r - \alpha_l} \\
    \text{is\_free}_i(\textbf{x}) &= d < \left( (1 - \delta) S_i[\alpha_l] + \delta S_i[\alpha_r] \right) 
\end{split}
\end{equation}

We then accumulate several timesteps $i\in [0, N-1]$ to build a ground-truth occupancy map.
In practice, we consider $N=20$ timesteps.
\begin{equation}
    \text{occ}(\textbf{x}) = \neg\left(\bigvee_{i \in [0, N]}\text{is\_free}_i(\textbf{x})\right)
\end{equation}
Similarly, we determine visibility by only considering the Lidar scan corresponding to the input frame:
\begin{equation}
    \text{vis}(\textbf{x}) = \neg \text{is\_free}_0(\textbf{x})
\end{equation}

Based on these functions, we can compute the final metric results.
We consider a point $\textbf{x}$ to be occupied, if the predicted density is over a threshold: $\sigma_\textbf{x} > 0.5$.
Let $\textbf{x}_i, i \in [1, N_\text{pts}]$ be points we sample from the camera frustum.
Let $X_{\neg\text{vis}} = \{i \in [1, N_\text{pts}] | \neg \text{vis}(\textbf{x}_i)\}$ be the subset of points that are invisible, and $X_{\neg\text{vis}\wedge \neg\text{occ}} = \{i \in [1, N_\text{pts}] | \neg \text{vis}(\textbf{x}_i) \wedge \neg \text{occ}(\textbf{x}_i)\}$ be the subset of points that are invisible \textit{and} empty.
\begin{equation}
    O_\text{acc} = \frac{1}{N_\text{pts}} \sum_{i=1}^{N_\text{pts}}\left(\text{occ}(\textbf{x}) == \left(\sigma_\textbf{x} > 0.5\right)\right)
\end{equation}
\begin{equation}
    \text{IE}_\text{acc} = \frac{1}{|X_{\neg\text{vis}}|} \sum_{i \in X_{\neg\text{vis}}}^{}\left(\text{occ}(\textbf{x}) == \left(\sigma_\textbf{x} > 0.5\right)\right)
\end{equation}
\begin{equation}
    \text{IE}_\text{rec} = \frac{1}{|X_{\neg\text{vis}\wedge \neg\text{occ}}|} \sum_{i \in X_{\neg\text{vis}\wedge \neg\text{occ}}}^{}\left(\sigma_\textbf{x} < 0.5\right)
\end{equation}

\begin{figure}
    \centering
    \includegraphics[trim={0cm .3cm 0cm 0cm},width=\linewidth]{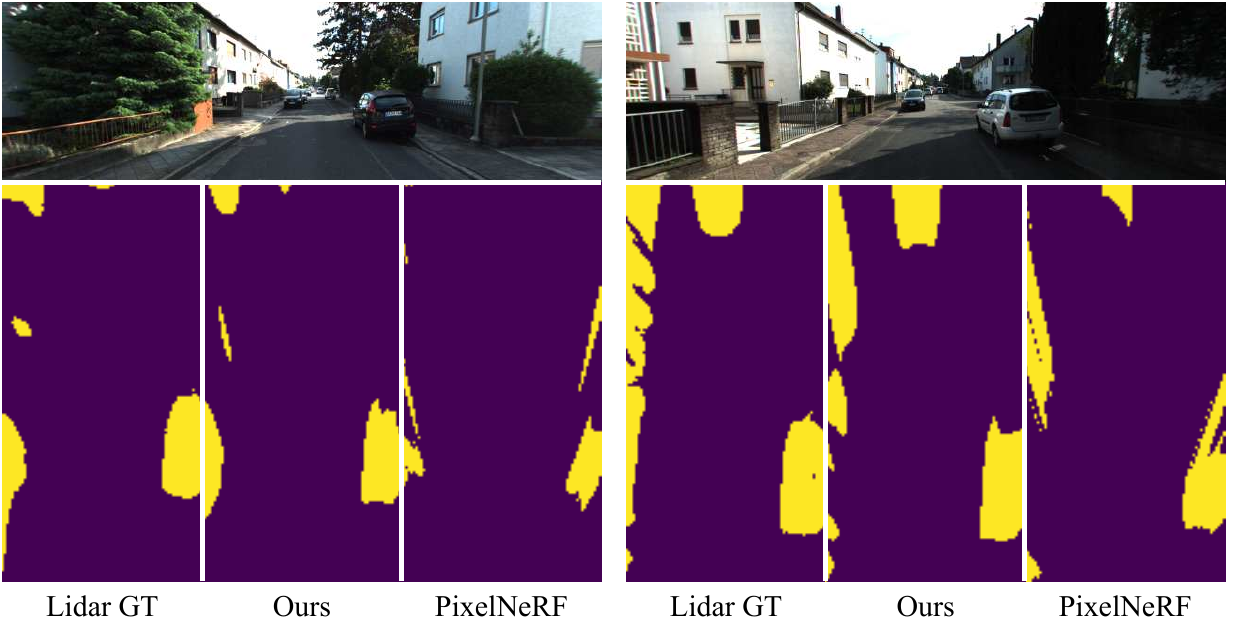}
    \caption{\textbf{Occupancy Metric on KITTI-360.} Visualization of the 1.\ occupancy ground-truth accumulated from 20 Lidar scans, 2.\ predicted occupancy map by our model, and 3.\ predicted occupancy map by PixelNeRF \cite{yu2021pixelnerf}.}
    \label{fig:lidar_metric}
\end{figure}

We sample 2720 points in total, uniformly spaced from a cuboid with dimensions $x = [-4m, 4m], y=[0m, 1m], z=[3m, 20m]$ (y-axis facing downward).
This means that all points are just above the surface of the street.
\cref{fig:lidar_metric} shows examples of the evaluation for two samples.
Evaluation code will be included in the code release. 

\section{Additional Considerations}

In this section, we discuss hypotheses on the working principles of our proposed approach, for which we do not have immediate experimental results.

\paragraph{Training Stability.}

Contrary to many NeRF-based methods, we find that our proposed approach offers stable training and that it is not overly sensitive to changes in hyperparameters.
We hypothesize, that this is due to the nature of color sampling, which shares similarities with classical stereo matching.
When casting a ray and sampling the color from a frame, the sampling positions will lie on the epipolar line.
The best match on this epipolar line, which would be the desired correspondence point in stereo matching, will give the smallest loss and a clear training signal.
This is even the case when sampling color from a single or very few frames.
In contrast, with a NeRF formulation, the color gets learned when multiple rays with the same color go through the same area in space.
Therefore, here we require many views to give a meaningful signal.

\paragraph{Reconstruction Quality.}
One of the key advantages of NeRF-based methods is that they offer a great way to aggregate the information from many frames that see the same areas of a scene.
In our formulation, color is only aggregated through the $\min$ operation in the loss term.
In a setting with many views, NeRF would clearly provide better reconstruction quality than a density field with color sampling.

However, in settings, where there are only few view scenes per scene available, most areas in the scene have very limited view coverage.
This means, that the aggregation aspect of NeRFs becomes much less relevant and the "visual expressiveness" of NeRFs and density fields with color sampling converge.

\section{Visualizations}

Assets for Fig.\ 2 were taken from Blendswap\footnote{\href{https://blendswap.com/blend/18686}{https://blendswap.com/blend/18686}}\footnote{\href{https://blendswap.com/blend/13698}{https://blendswap.com/blend/13698}} under the CC-BY license.

\begin{table*}
\centering
\footnotesize
\begin{tabular}{lccccccccc}
\toprule
\textit{Model} & Volumetric & Split & Abs Rel & Sq Rel & RMSE & $\text{RMSE}_{\log}$ & $\alpha < 1.25$ & $\alpha < 1.25^2$ & $\alpha < 1.25^3$ \\
\midrule
PixelNeRF \cite{yu2021pixelnerf} & \cmark & \multirow{7}{*}{Eigen \cite{eigen2014depth}} & 0.130 & 1.241 & 5.134 & 0.220 & 0.845 & 0.943 & 0.974 \\
EPC++ \cite{luo2019every}        & \xmark &  & 0.128 & 1.132 & 5.585 & 0.209 & 0.831 & 0.945 & 0.979 \\
MonoDepth 2 \cite{godard2019digging} & \xmark &  & 0.106 & 0.818 & 4.750 & 0.196 & 0.874 & 0.957 & 0.975 \\
PackNet \cite{guizilini20203d}   & \xmark &  & 0.111 & 0.785 & 4.601 & 0.189 & 0.878 & 0.960 & \underline{0.982} \\
DepthHint \cite{watson2019self}  & \xmark &  & 0.105 & 0.769 & 4.627 & 0.189 & 0.875 & 0.959 & \underline{0.982} \\
FeatDepth \cite{shu2020feature}  & \xmark &  & \underline{0.099} & \underline{0.697} & 4.427 & \underline{0.184} & \underline{0.889} & \underline{0.963} & \underline{0.982} \\
DevNet  \cite{zhou2022devnet}    & (\cmark) &  & \textbf{0.095} & \textbf{0.671} & \textbf{4.365} & \textbf{0.174} & \textbf{0.895} & \textbf{0.970} & \textbf{0.988} \\
\midrule
\textbf{Ours} & \cmark &  & 0.102 & 0.751 & \underline{4.407} & 0.188 & 0.882 & 0.961 & \underline{0.982} \\ \midrule \midrule
MINE \cite{li2021mine} & \cmark & \multirow{2}{*}{Tulsiani \cite{tulsiani2018layer}} & 0.137 & 1.993 & 6.592 & 0.250 & 0.839 & 0.940 & 0.971 \\
\textbf{Ours} & \cmark &  & \textbf{0.132} & \textbf{1.936} & \textbf{6.104} & \textbf{0.235} & \textbf{0.873} & \textbf{0.951} & \textbf{0.974}\\
\bottomrule
\end{tabular}
\caption{
\textbf{Depth Prediction on KITTI.} 
While our goal is full volumetric scene understanding, we compare to state-of-the-art self-supervised depth estimation method. 
Our approach achieves competitive performance while clearly improving over other volumetric approaches like PixelNeRF \cite{yu2021pixelnerf} and MINE \cite{li2021mine}.
DevNet \cite{zhou2022devnet} performs better, but does not show any results of their volume. 
\label{tab:depth_prediction_full}
}
\end{table*}

\end{document}